\documentclass[10pt,twocolumn,letterpaper]{article}

\usepackage{cvpr}

\definecolor{navyblue}{HTML}{0071BC}
\definecolor{hotpink}{HTML}{FF0080}

\definecolor{oai-white}{HTML}{FFFFFF}
\definecolor{oai-black}{HTML}{000000}
\definecolor{oai-red}{HTML}{FF4500}
\definecolor{oai-green}{HTML}{51DA4C}
\definecolor{oai-blue}{HTML}{0000FF}
\definecolor{oai-yellow}{HTML}{FFF639}
\definecolor{oai-magenta}{HTML}{FF45FF}
\definecolor{oai-cyan}{HTML}{00FFFF}
\definecolor{oai-orange}{HTML}{FE7600}
\definecolor{oai-violet}{HTML}{8A2BE2}
\definecolor{oai-brown}{HTML}{A0522D}
\definecolor{oai-green-050}{HTML}{F4FFF4}
\definecolor{oai-green-100}{HTML}{E9FFE8}
\definecolor{oai-green-200}{HTML}{D9FFD8}
\definecolor{oai-green-300}{HTML}{C9FFC7}
\definecolor{oai-green-400}{HTML}{A6FFA3}
\definecolor{oai-green-500}{HTML}{7CF178}
\definecolor{oai-green-600}{HTML}{51DA4C}
\definecolor{oai-green-700}{HTML}{3FA93B}
\definecolor{oai-green-800}{HTML}{2D712A}
\definecolor{oai-green-900}{HTML}{193718}
\definecolor{oai-gray-000}{HTML}{FFFFFF}
\definecolor{oai-gray-100}{HTML}{FAFAFA}
\definecolor{oai-gray-200}{HTML}{F5F5F5}
\definecolor{oai-gray-300}{HTML}{E5E5E5}
\definecolor{oai-gray-400}{HTML}{FFB7A4}
\definecolor{oai-gray-500}{HTML}{CDCDCD}
\definecolor{oai-gray-600}{HTML}{A8A8A8}
\definecolor{oai-gray-700}{HTML}{747474}
\definecolor{oai-gray-800}{HTML}{393939}
\definecolor{oai-gray-900}{HTML}{000000}
\definecolor{visual}{HTML}{A50E0E}       
\definecolor{linguistic}{HTML}{174EA6}   
\definecolor{relational}{HTML}{E37400}   
\definecolor{egocentric}{HTML}{0D652D}  
\usepackage[pagebackref,breaklinks,colorlinks,allcolors=navyblue]{hyperref}

\usepackage{graphicx}
\usepackage{amssymb}
\usepackage{adjustbox}
\usepackage{colortbl}
\usepackage{xcolor}
\usepackage{multirow}
\usepackage{array}
\usepackage{makecell}
\usepackage{bbm}
\usepackage{collcell,xfp}
\usepackage{pgf}
\usepackage{tikz}
\usepackage[most]{tcolorbox}
\usepackage{csquotes}
\usepackage[noorphans,vskip=1em,leftmargin=1em]{quoting}
\usepackage{pifont}
\usepackage{enumitem} 
\usepackage{tcolorbox} 
\usepackage{forest}
\usepackage{wrapfig}
\usepackage{caption}
\usepackage{subcaption}
\usepackage{longtable}
\usepackage[T1]{fontenc}

\colorlet{mapcolor}{ForestGreen}

\title{Thinking in Space: How Multimodal Large Language Models \\ See, Remember, and Recall Spaces}

\newcommand{\fmtnum}[1]{%
  \ifnum\fpeval{#1 < 0} = 1
    \textcolor{red}{$#1$}%
  \else
      \textcolor{green}{$#1$}%
  \fi
}

\definecolor{linkblue}{rgb}{0.1, 0.5, 0.7}

\newcommand{\infobox}[1]{
    \vspace{-0.18cm}
    \begin{tcolorbox}[
        colback=white!90!gray,     
        colframe=teal!60!black,   
        arc=5pt,                   
        boxsep=5pt,                 
        left=5pt,                  
        right=10pt,                 
        top=2pt,                   
        bottom=3pt,                
        boxrule=0.8pt,              
        drop shadow=gray!50!white, 
        enhanced jigsaw             
    ]
    \vspace{-0.1cm}
         \textit{#1}
    \vspace{-0.2cm}
    \end{tcolorbox}
    \vspace{-0.15cm}
}

\usepackage{fontawesome}
\newcommand{\bench}{\texttt{VSI-Bench}\xspace}
\newcommand{\benchtiny}{\texttt{VSI-Bench} (tiny)\xspace}

\newcommand{\geminipro}{Gemini-1.5 Pro}

\newcommand{\worldwideweb}{\raisebox{-1.5pt}{\includegraphics[height=1.05em]{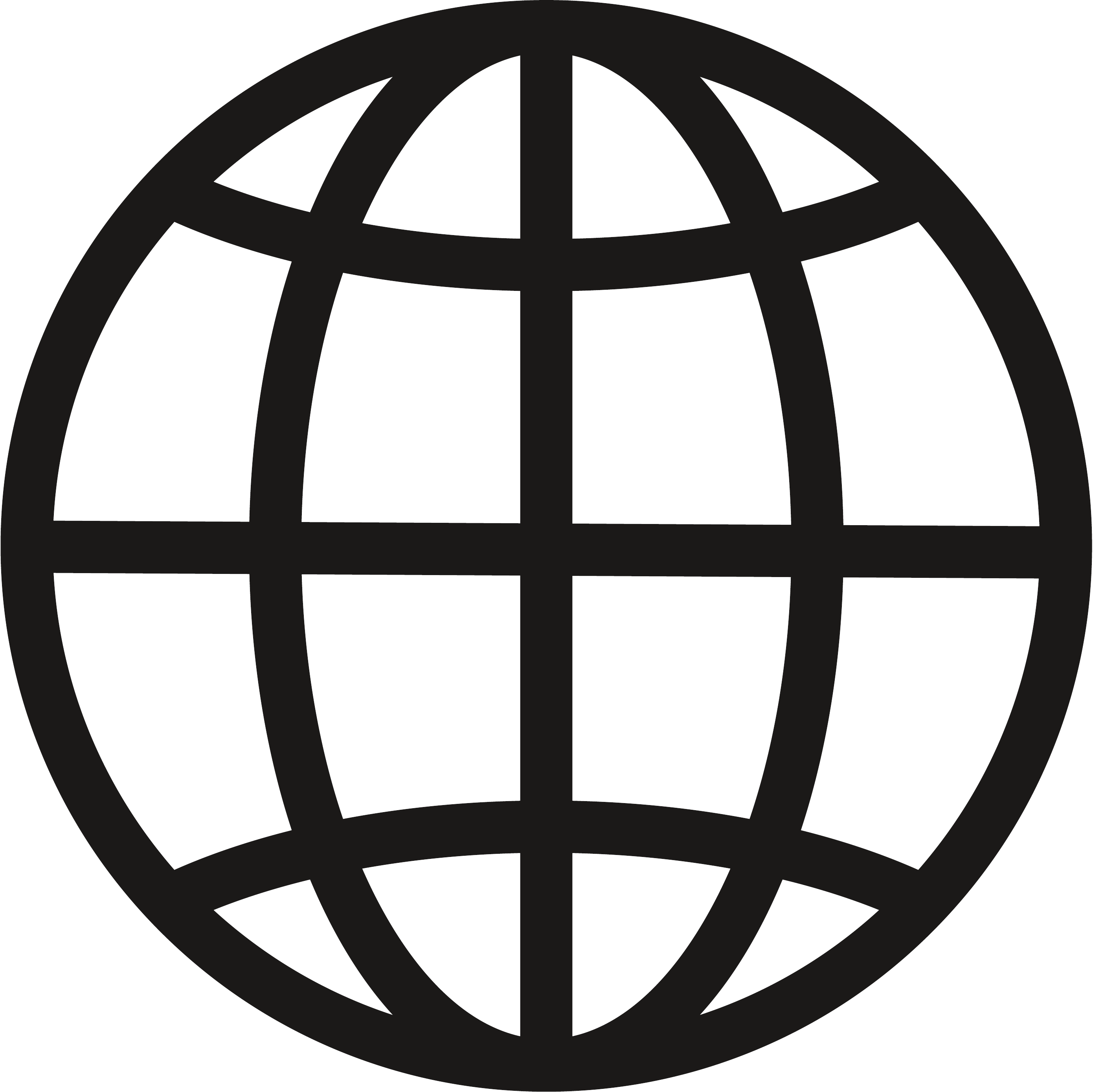}}\xspace}
\newcommand{\github}{\raisebox{-1.5pt}{\includegraphics[height=1.05em]{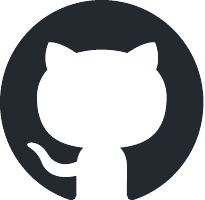}}\xspace}
\newcommand{\huggingface}{\raisebox{-1.5pt}{\includegraphics[height=1.05em]{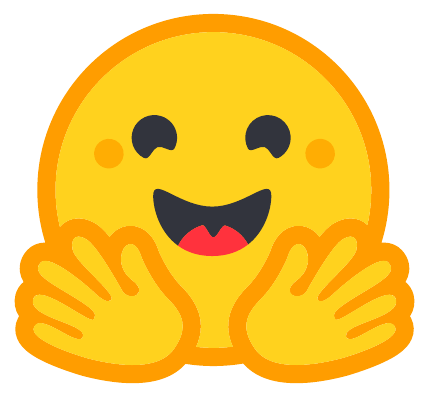}}\xspace}

\author{
Jihan Yang\textsuperscript{1}\footnotemark[1] \quad
Shusheng Yang\textsuperscript{1}$^*$ \quad
Anjali W. Gupta\textsuperscript{1}$^*$ \quad
Rilyn Han\textsuperscript{2}$^*$ \quad
Li Fei-Fei\textsuperscript{3} \quad
Saining Xie\textsuperscript{1} \vspace{.2em} \\
\textsuperscript{1}New York University \quad  \textsuperscript{2}Yale University \quad
\textsuperscript{3}Stanford University \\
\\
{\worldwideweb \href{https://vision-x-nyu.github.io/thinking-in-space.github.io/}{{\text{Project Page}}}} \quad \quad {\github \href{https://github.com/vision-x-nyu/thinking-in-space}{{\text{Evaluation Code}}}}
\quad \quad
{\huggingface \href{https://huggingface.co/datasets/nyu-visionx/VSI-Bench}{{\text{VSI-Bench}}}}
\vspace{0.2cm}
}

\begin{document}

\twocolumn[{
    \renewcommand\twocolumn[1][]{#1}
    \maketitle
    \vspace*{-0.4in}
    \centering
    \captionsetup{type=figure}
    % \hspace*{-1cm}
    \includegraphics[width=1\textwidth]{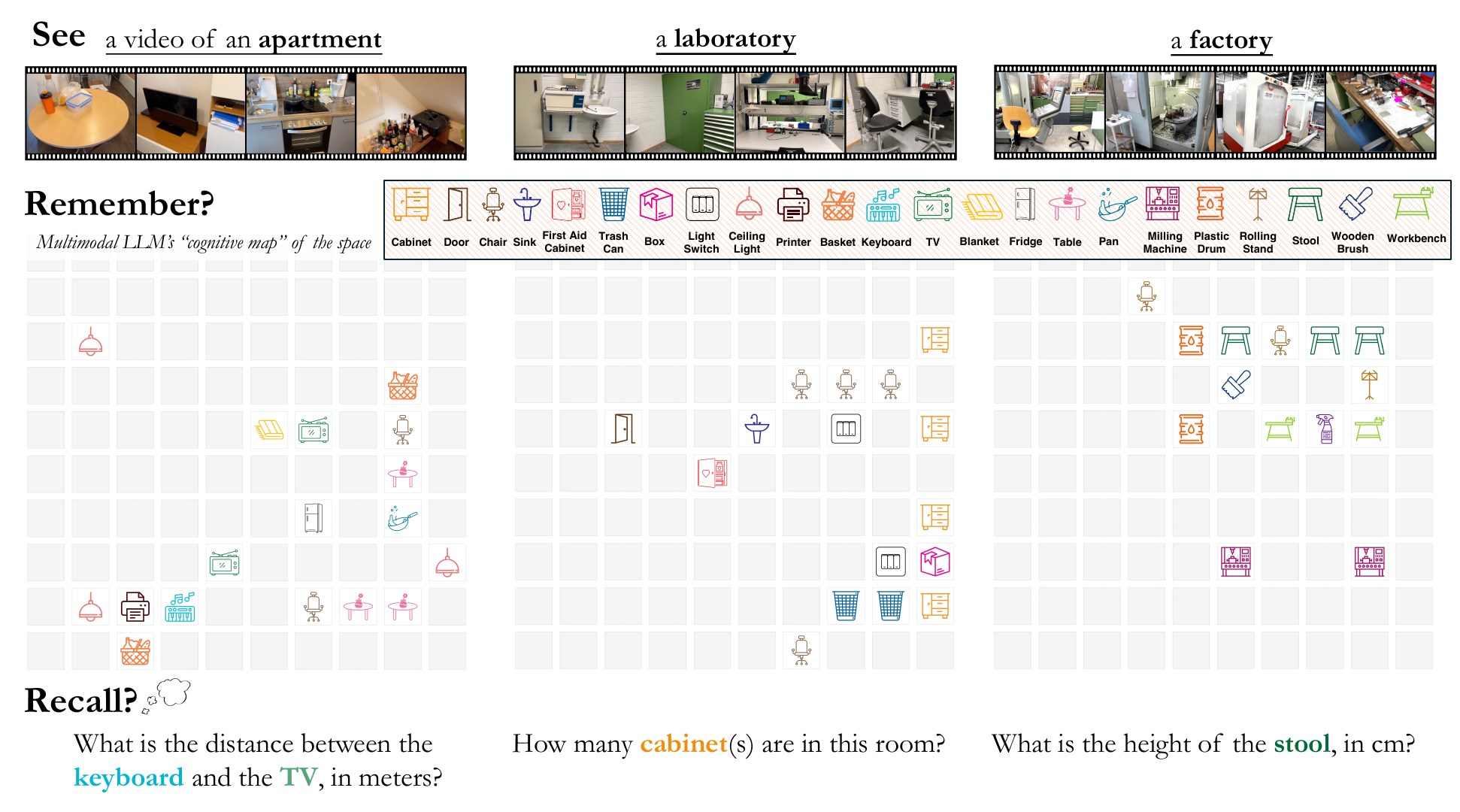}
    \vspace{-0.9cm}
    \caption{Whether at home, in the workplace, or elsewhere, the ability to perceive a space, remember its layout, and retrieve this spatial information to answer questions on demand is a key aspect of visual-spatial intelligence. Recent Multimodal LLMs can understand general videos, but can they ``think spatially'' when presented with a video recording of an environment? Can they build an accurate, implicit ``\emph{cognitive map}'' that allows them to answer questions about a space? What are the strengths and limitations of using MLLMs to enhance spatial intelligence? We dig into these questions by setting up video data for MLLMs to watch, building a VQA benchmark to check their recall, and examining what the MLLMs actually remember and understand.}
    \label{fig:teaser}
    \vspace{0.3cm}
}]

\renewcommand{\thefootnote}{\fnsymbol{footnote}}
\footnotetext[1]{Equal contribution.}

\begin{abstract}
\noindent Humans possess the visual-spatial intelligence to remember spaces from sequential visual observations. However, can Multimodal Large Language Models (MLLMs) trained on million-scale video datasets also ``think in space'' from videos? 
We present a novel video-based visual-spatial intelligence benchmark (VSI-Bench) of over 5,000 question-answer pairs, and find that MLLMs exhibit competitive\textemdash though subhuman\textemdash visual-spatial intelligence.
We probe models to express how they think in space both linguistically and visually and find that while spatial reasoning capabilities remain the primary bottleneck for MLLMs to reach higher benchmark performance, local world models and spatial awareness do emerge within these models. 
Notably, prevailing linguistic reasoning techniques (\eg, chain-of-thought, self-consistency, tree-of-thoughts) fail to improve performance, whereas explicitly generating cognitive maps during question-answering enhances MLLMs' spatial distance ability.

\end{abstract}

\vspace{-0.3cm}
\section{Introduction}
\vspace{-0.2cm}
When shopping for furniture, we often try to recall our living room to imagine if a desired cabinet will fit. Estimating distances is difficult, yet after even a single viewing, humans can mentally reconstruct spaces, recalling objects in a room, their positions, and sizes. We live in a sensory-rich 3D world where visual signals surround and ground us, allowing us to perceive, understand, and interact with it.

Visual-spatial intelligence entails perceiving and mentally manipulating spatial relationships \cite{gardner1983frames}; it requires myriad capabilities, including relational reasoning and the ability to transform between egocentric and allocentric perspectives (\cref{sec:vsi}). While Large Language Models (LLMs)~\cite{naveed2024comprehensiveoverviewlargelanguage, wei2022emergent, beguvs2023large, zhang-etal-2024-unveiling-linguistic, Kassner2023LanguageMW,radford2018improving,radford2019language,brown2020language,touvron2023llama,touvron2023llama2,bai2023qwen,team2023gemini} have advanced linguistic intelligence, visual-spatial intelligence remains under-explored, despite its relevance to robotics~\cite{driess2023palm,brohan2022rt,brohan2023rt,o2023open}, autonomous driving~\cite{tian2024drivevlm}, and AR/VR~\cite{chandrasegaran2024hourvideo,grauman2022ego4d,mangalam2023egoschema}.

Multimodal Large Language Models (MLLMs)~\cite{hurst2024gpto,team2024gemini,liu2024visual,alayrac2022flamingo,li2023blip2,liu2024visual,bai2023qwenvl,chen2024internvl}, which integrate language and vision, exhibit strong capacities to think and reason in open-ended dialog and practical tasks like web agents~\cite{jimenez2024swebench,gur2023real,huang2022language,driess2023palm}. 
To advance this intelligence in the visual-spatial realm, we introduce \bench, a video-based benchmark featuring over 5,000 question-answer pairs across nearly 290 real indoor-scene videos (\cref{sec:benchmark}). Video data, by capturing continuous, temporal input, both parallels how we observe the world and enables richer spatial understanding and reasoning than static images. Evaluating open- and closed-source models on \bench reveals that even though a large performance gap exists between models and humans, MLLMs exhibit emerging visual-spatial intelligence despite the challenges of video understanding, textual understanding, and spatial reasoning (\cref{sec:evaluation}). 

To analyze model behavior and inspired by dual-coding theory \cite{clark1991dual}, which posits that linguistic and visual processing are distinct yet complementary, we prompt selected models for self-explanations (linguistic) and cognitive maps (visual). Analyzing the self-explanations reveals that spatial reasoning, as compared to visual perception, linguistic intelligence, or temporal processing, is the main factor behind weak performance on \bench (\cref{sec:think_linguistic}). ``\emph{Cognitive maps}'', which represent internal layouts of environments~\cite{tolman1948cognitive,Newcombe2024Spatial}, allow us to evaluate MLLMs' implicit spatial world models and find that MLLMs build strong local models but weak global ones (\cref{sec:think_visual}). Furthermore, standard linguistic reasoning techniques fail to enhance performance on our benchmark. However, explicitly generating and using cognitive maps improves spatial distance question-answering.

\begin{figure}[t]
\centering
        \includegraphics[clip,trim=0cm 0cm 0cm 0cm,width=0.95\linewidth]{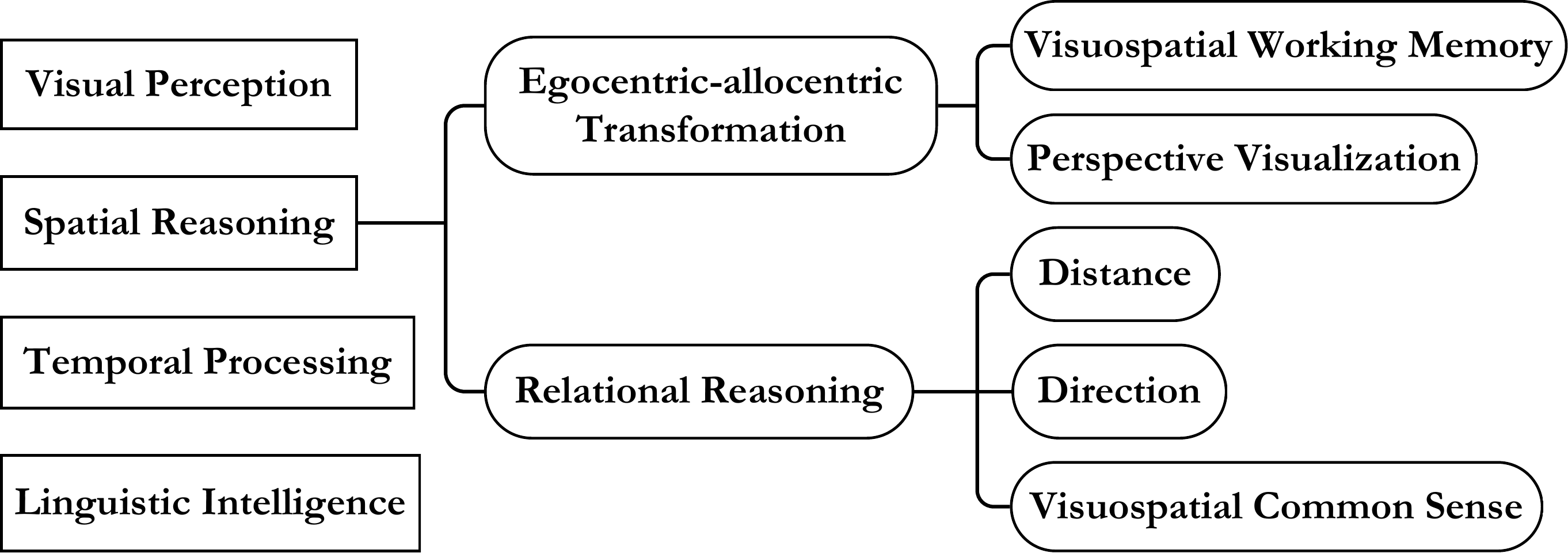}
     \caption{A taxonomy of \textbf{visual-spatial intelligence} capabilities.}
    \label{fig:vsi-taxonomy}
    \vspace{-6mm}
\end{figure}

Expressing visual-spatial intelligence is difficult (and often piecemeal), even for humans \cite{gardner1983frames}.
With this work, we aim to encourage the community to explore grounding frontier models with visual-spatial intelligence and to pave and illuminate this direction.

\vspace{-0.2cm}
\section{Visual-Spatial Intelligence}
\vspace{-0.1cm}
\label{sec:vsi}
We discuss preliminaries and scope visual-spatial intelligence to provide context and a framework for later analysis.

\vspace{0.05cm}
\noindent\textbf{Term Use.} We use ``intelligence'' rather than ``cognition'' as it is broader, and ``spatial cognition'' is a branch of cognitive psychology \cite{waller2013handbook}. We prefix spatial intelligence in our work with ``visual'', as spatial intelligence exists irrespective of sensory modality  (\eg, a blind person can perceive space through other senses) \cite{gardner1983frames}. Given our focus on video input, we discuss \textit{visual}-spatial intelligence. 

\vspace{0.05cm}
\noindent\textbf{Investigation Scope.} 
While classic spatial intelligence tests also include pen-paper tasks like the Mental Rotation Test~\cite{shepard1988mental}, our focus is on visual-spatial intelligence as it applies to real-world environments, particularly in common spaces like homes, offices, and factories. 

\noindent\textbf{Taxonomy.} We provide a taxonomy of capabilities potentially required for visual-spatial intelligence (\cref{fig:vsi-taxonomy}), based on cognitive psychology \cite{gardner1983frames, meneghetti2022individual, Newcombe2024Spatial, chabris_jerde_woolley_gerbasi_schuldt_wai_bennett_hackman_kosslyn_2023} and human experience with our benchmark tasks in \cref{sec:benchmark}. Visual perception, linguistic intelligence, temporal processing, and spatial reasoning are the four areas needed in \bench. For example, \cite{chabris_jerde_woolley_gerbasi_schuldt_wai_bennett_hackman_kosslyn_2023} shows that visual object and spatial processing are neurally distinct, which motivates ``visual perception'' and ``spatial reasoning'' as separate areas. 
We break spatial reasoning into two broad capabilities: relational reasoning and egocentric-allocentric transformation. 

\begin{figure*}[t]
\centering
\vspace{-5mm}
        \includegraphics[clip,trim=0cm 0cm 0cm 0cm,width=1.0\linewidth]{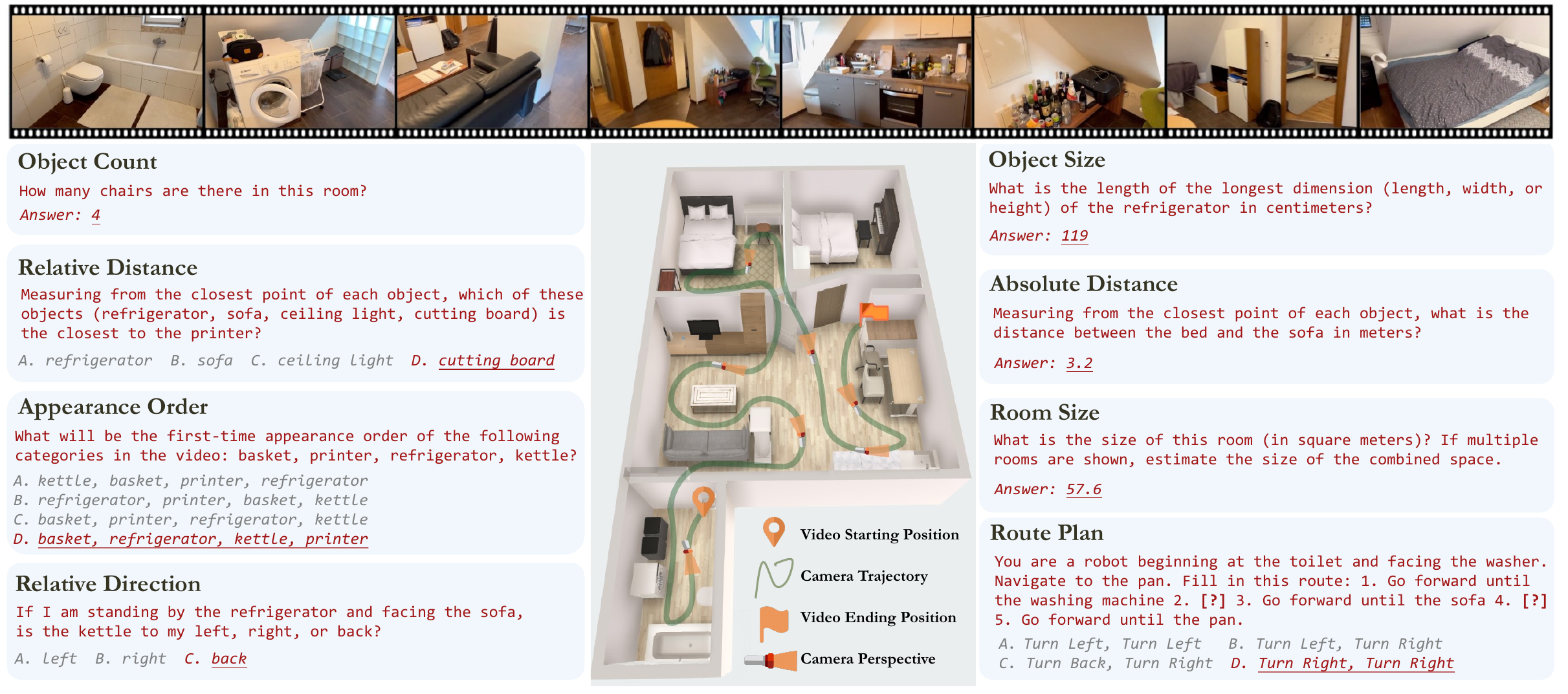}
\vspace{-7mm}
    \caption{\textbf{Tasks demonstration of \bench}. Note: the questions above are simplified slightly for clarity and brevity.}
    \label{fig:task_demonstration}
    \vspace{-4mm}
\end{figure*}

\textit{Relational reasoning} is the ability to identify, via distance and direction, relationships between objects. It also encompasses reasoning about distance between objects by relying on visuospatial common sense about the sizes of other objects. For example, knowing a standard beverage can is approximately 12 cm tall, humans can estimate other object sizes by comparing visual proportions. 

\textit{Egocentric-allocentric transformation} involves shifting between a self-centered (egocentric) view and an environment-centered (allocentric) one. In our setting, each egocentric video frame maps to allocentric object positions and camera trajectory. When humans observe a space, they convert egocentric perceptions into an allocentric mental map, enabling perspective-taking from various viewpoints\textemdash essential for tasks like relative direction or route planning. This transformation relies on visualizing new perspectives and on visuospatial working memory~\cite{baddeley1992working}, the ability to hold and manipulate spatial information, say by updating object positions from new egocentric input \cite{mcafoose2009exploring, dehn2011working}.

Every task in \bench requires perceptual, linguistic, and temporal abilities and varying degrees of spatial reasoning. For example, egocentric-allocentric transformation is much more important for a task like route planning than object size estimation. These factors provide some context for the complexity of visual-spatial intelligence.

\section{\bench}
\vspace{-0.1cm}
\label{sec:benchmark}
\subsection{Overview}
\vspace{-0.1cm}
\label{sec:benchmark_overview}
We introduce \bench to quantitatively evaluate the visual-spatial intelligence of MLLMs from egocentric video. \bench comprises over 5,000 question-answer pairs derived from 288 real videos. These videos are sourced from the validation sets of the public indoor 3D scene reconstruction datasets ScanNet~\cite{dai2017scannet}, ScanNet++~\cite{yeshwanth2023scannet++}, and ARKitScenes~\cite{dehghan2021arkitscenes} and represent diverse environments\textemdash including residential spaces, professional settings (\eg, offices, labs), and industrial spaces (\eg, factories)\textemdash and multiple geographic regions. Repurposing these existing 3D reconstruction and understanding datasets offers accurate object-level annotations which we use in question generation and could enable future study into the connection between MLLMs and 3D reconstruction. \bench is high-quality, having been iteratively reviewed to minimize question ambiguity and to remove incorrect annotations propagated from the source datasets. 

\begin{figure*}[tbh]
\centering
\vspace{-0.4cm}
        \includegraphics[clip,trim=0cm 0cm 0cm 0cm,width=1\linewidth]{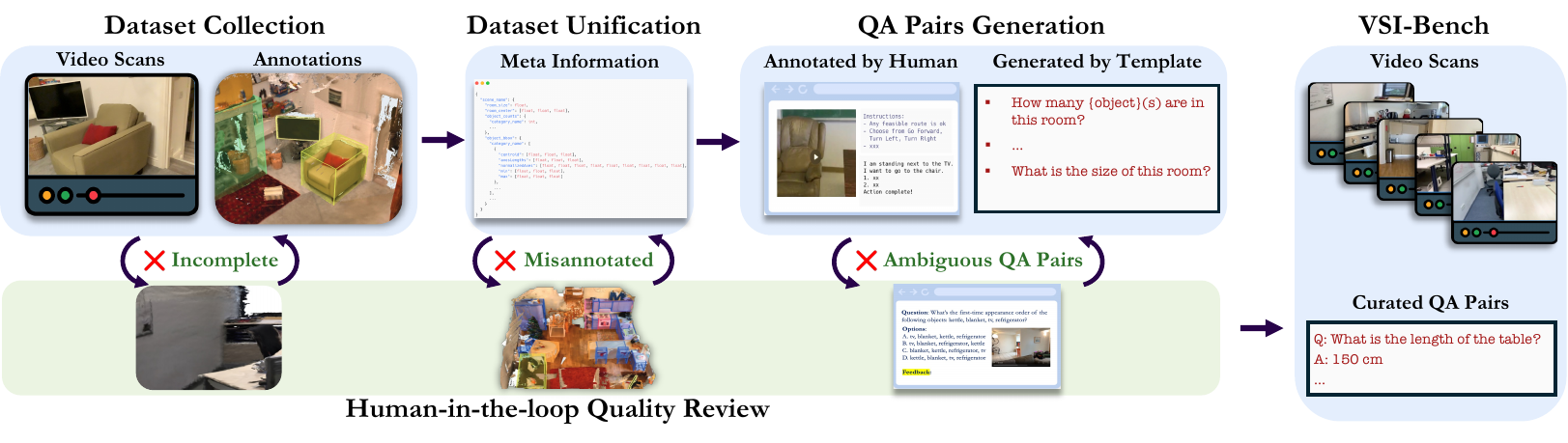}
        \vspace{-0.9cm}
\caption{\textbf{Benchmark curation pipeline.} The pipeline first unifies diverse datasets into a standardized format and semantic space for consistent processing. QA pairs are then generated through both human annotation and question templates. To ensure quality, human verification is implemented at all key stages for filtering low-quality videos, annotations, and ambiguous QA pairs.}
\vspace{-4mm}
    \label{fig:benchmark_construct}
\end{figure*}

\begin{figure}[tbh]
\centering
        \includegraphics[clip,trim=0cm 0cm 0cm 0cm,width=0.9\linewidth]{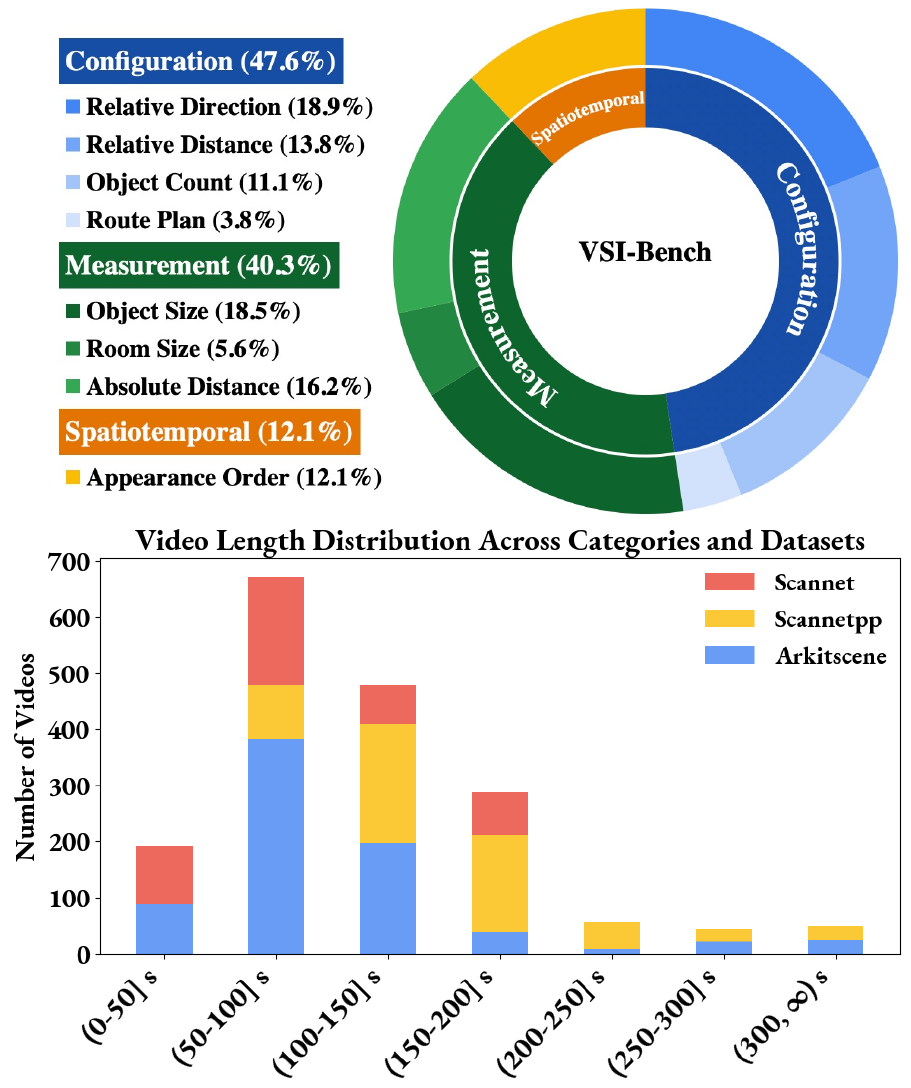}
    \vspace{-3mm}
    \caption{\textbf{Benchmark Statistics}. \textbf{Top}: The distribution of tasks across three main categories. \textbf{Bottom}: The video length statistic.}
    \label{fig:dataset-overview}
    \vspace{-6mm}
\end{figure}

\bench includes eight tasks of three types: \textit{configurational}, \textit{measurement estimation}, and \textit{spatiotemporal}. 
The configurational tasks (\textit{object count, relative distance, relative direction, route plan}) test a model’s understanding of the configuration of a space and are more intuitive for humans (see \cref{sec:evaluation} for comparison between MLLM and human performance). 
Measurement estimation (of \textit{object size, room size}, and \textit{absolute distance}) is of value to any embodied agent. While predicting a measurement exactly is very difficult, for both humans and models, a better sense of distance and other measurements is intuitively correlated with better visual-spatial intelligence and underpins a wide range of tasks that require spatial awareness, like interaction with objects and navigation. 
Spatiotemporal tasks like \textit{appearance order} test a model's memory of a space as seen in video. See \cref{fig:task_demonstration} for an overview of \bench tasks and \cref{fig:dataset-overview} for dataset statistics.

\subsection{Benchmark Construction}
We develop a sophisticated benchmark construction pipeline to effectively generate high-quality question-answer (QA) pairs at scale, as shown in
\cref{fig:benchmark_construct}.

\vspace{0.05cm}
\noindent\textbf{Data Collection and Unification.} We begin our dataset construction by standardizing various datasets into a unified meta-information structure, ensuring dataset-agnostic QA pair generation. Our benchmark aggregates existing 3D indoor scene understanding and reconstruction datasets: ScanNet~\cite{dai2017scannet}, ScanNet++~\cite{yeshwanth2023scannet++}, and ARKitScenes~\cite{dehghan2021arkitscenes}. These datasets provide high-fidelity video scans capable of space reconstruction, ensuring MLLMs can answer space-level questions with only video input. Additionally, their object-level 3D annotations facilitated our question generation. We parse the datasets into a unified meta-information format including object categories, bounding boxes, video specifications (resolution and frame rate), and more.

\vspace{0.05cm}
\noindent\textbf{Question-Answer Generation.} QA pairs are primarily auto-annotated using the meta-information and question templates; the \textit{route plan} task was human-annotated. We sophisticatedly design and refine the question template for each task and provide guidelines for human annotators. For more detailed design, see \cref{sec:supp_bench_construct}.

\begin{figure*}[ht!]
    \captionsetup{type=table}
    \vspace{-0.4cm}
    \centering
    \begin{minipage}{0.63\textwidth} 
    \centering
    \fontsize{4.6pt}{4.4pt}\selectfont
    \setlength\tabcolsep{3pt} 
    \renewcommand{\arraystretch}{1.2} 
    \scalebox{1.57}{
    \begin{tabular}{r|cc|cccccccc}
    & & &
    \rotatebox{75}{Obj. Count} &
    \rotatebox{75}{Abs. Dist.} &
    \rotatebox{75}{Obj. Size} & 
    \rotatebox{75}{Room Size} &
    \rotatebox{75}{Rel. Dist.} &
    \rotatebox{75}{Rel. Dir.} &
    \rotatebox{75}{Route Plan} &
    \rotatebox{75}{Appr. Order} \\
    Methods & Rank & Avg. & \multicolumn{4}{c}{\cellcolor{orange!10}Numerical Answer} & \multicolumn{4}{c}{\cellcolor{yellow!10}Multiple-Choice Answer} \\
    \hline
    \rowcolor{navyblue!5}
    \multicolumn{1}{l|}{\textcolor{black}{\textit{Baseline}}} & & & & & & & & & & \\
    Chance Level (Random) & - & - & - & - & - & - & 25.0 & 36.1 & 28.3 & 25.0 \\
    Chance Level (Frequency) & - & 34.0 & 62.1 & 32.0 & 29.9 & 33.1 & 25.1 & 47.9 & 28.4 & 25.2 \\
    \hline
    \rowcolor{navyblue!5}
    \multicolumn{1}{l|}{\textcolor{black}{\textit{\benchtiny Perf.}}} & & & & & & & & & & \\
    \textsuperscript{\dag}Human Level & - &79.2 &94.3 &47.0 &60.4 &45.9 &94.7 &95.8 &95.8 &100.0 \\
    \textsuperscript{\dag}Gemini-1.5 Flash & - & 45.7 & 50.8 & 33.6 & 56.5 & 45.2 & 48.0 & 39.8 & 32.7 & 59.2 \\
    \textsuperscript{\dag}Gemini-1.5 Pro & - & 48.8 & 49.6 & 28.8 & 58.6 & 49.4 & 46.0 & 48.1 & 42.0 & 68.0 \\
    \textsuperscript{\dag}Gemini-2.0 Flash & - & 45.4 & 52.4 & 30.6 & 66.7 & 31.8 & 56.0 & 46.3 & 24.5 & 55.1 \\
    \hline
    \rowcolor{navyblue!5}
    \multicolumn{1}{l|}{\textcolor{black}{\textit{Proprietary Models (API)}}} & & & & & & & & & & \\
    GPT-4o & \cellcolor{oai-green-200}{3} & 34.0 & 46.2 & 5.3 & 43.8 & 38.2 & 37.0 & 41.3 & 31.5 & 28.5 \\
    Gemini-1.5 Flash & \cellcolor{oai-green-400}{2} & 42.1 & 49.8 & 30.8 & 53.5 & \cellcolor{oai-gray-600}{54.4} & 37.7 & 41.0 & 31.5 & 37.8 \\
    Gemini-1.5 Pro & \cellcolor{oai-green-600}{1} & 45.4 & \cellcolor{oai-gray-600}{56.2} & \cellcolor{oai-gray-600}{30.9} & \cellcolor{oai-gray-600}{64.1} & 43.6 & \cellcolor{oai-gray-600}{51.3} & \cellcolor{oai-gray-600}{46.3} & \cellcolor{oai-gray-600}{36.0} & 34.6 \\
    \hline
    \rowcolor{navyblue!5}
    \multicolumn{1}{l|}{\textcolor{black}{\textit{Open-source Models}}} & & & & & & & & & & \\
    % InternVL2-2B & 11 & 27.4 & 21.8 & 24.9 & 22.0 & 35.0 & 33.8 & \cellcolor{oai-gray-300}{44.2} & 30.5 & 7.1 \\
    InternVL2-2B & 11 & 26.5 & 25.7 & 24.0 & 20.0 & 29.2 & 32.1 & \cellcolor{oai-gray-300}{44.1} & 30.4 & 6.3 \\
    % InternVL2-8B & 5 & 34.6 & 23.1 & \cellcolor{oai-gray-300}{28.7} & 48.2 & \cellcolor{oai-gray-300}{39.8} & 36.7 & 30.7 & 29.9 & 39.6 \\
    InternVL2-8B & \cellcolor{oai-green-200}{3} & 37.5 & 31.3 & \cellcolor{oai-gray-300}{29.0} & 48.9 & \cellcolor{oai-gray-300}{44.2} & 38.0 & 33.4 & 28.9 & 46.4 \\
    % InternVL2-40B & \cellcolor{oai-green-200}{3} & 36.0 & 34.9 & 26.9 & 46.5 & 31.8 & 42.1 & 32.2 & 34.0 & 39.6 \\
    InternVL2-40B & 4 & 37.0 & 41.3 & 26.2 & 48.2 & 27.5 & 47.6 & 32.7 & 27.8 & 44.7 \\
    LongVILA-8B & 12 & 21.6 & 29.1 & 9.1 & 16.7 & 0.0 & 29.6 & 30.7 & 32.5 & 25.5 \\
    VILA-1.5-8B & 9 & 28.9 & 17.4 & 21.8 & 50.3 & 18.8 & 32.1 & 34.8 & 31.0 & 24.8 \\
    VILA-1.5-40B & 7 & 31.2 & 22.4 & 24.8 & 48.7 & 22.7 & 40.5 & 25.7 & 31.5 & 32.9 \\
    LongVA-7B & 8 & 29.2 & 38.0 & 16.6 & 38.9 & 22.2 & 33.1 & 43.3 & 25.4 & 15.7 \\
    LLaVA-Video-7B & 5 & 35.6 & 48.5 & 14.0 & 47.8 & 24.2 & \cellcolor{oai-gray-300}{43.5} & 42.4 & 34.0 & 30.6 \\
    LLaVA-Video-72B & \cellcolor{oai-green-600}{1} & 40.9 & \cellcolor{oai-gray-300}{48.9} & 22.8 & 57.4 & 35.3 & 42.4 & 36.7 & \cellcolor{oai-gray-300}{35.0} & \cellcolor{oai-gray-600}{48.6} \\
    LLaVA-OneVision-0.5B & 10 & 28.0 & 46.1 & 28.4 & 15.4 & 28.3 & 28.9 & 36.9 & 34.5 & 5.8 \\
    LLaVA-OneVision-7B & 6 & 32.4 & 47.7 & 20.2 & 47.4 & 12.3 & 42.5 & 35.2 & 29.4 & 24.4 \\
    LLaVA-OneVision-72B & \cellcolor{oai-green-400}{2} & 40.2 & 43.5 & 23.9 & \cellcolor{oai-gray-300}{57.6} & 37.5 & 42.5 & 39.9 & 32.5 & 44.6 \\
    \end{tabular}
} 
    \end{minipage}
    \hfill
    \begin{minipage}[c]{0.35\textwidth}
        \centering
        \includegraphics[width=\textwidth]{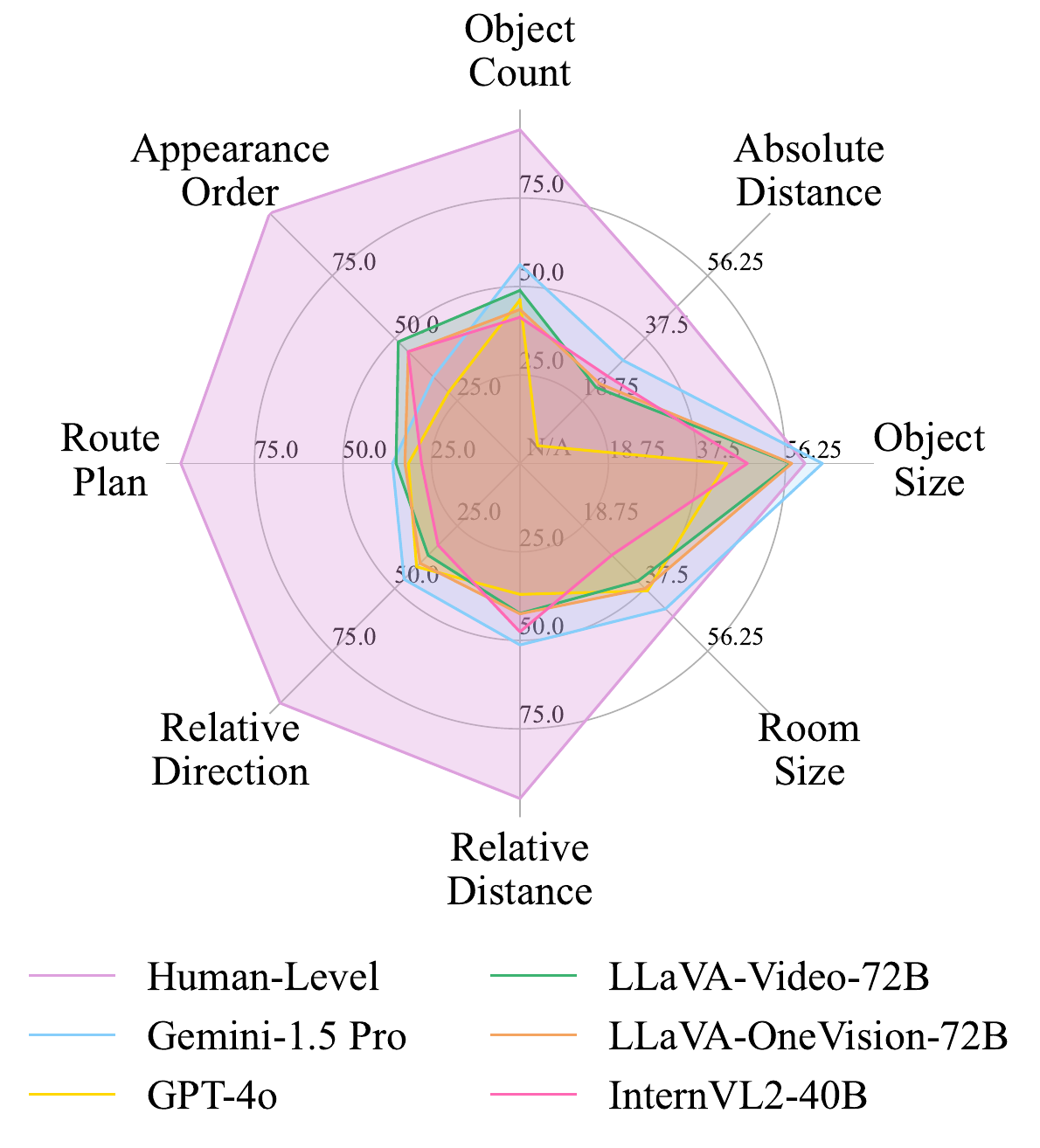} 
    \end{minipage}
    \vspace{-0.2cm}
    \caption{\textbf{Evaluation on \bench.} \textbf{Left:} \colorbox{oai-gray-600}{Dark gray} indicates the best result among all models and \colorbox{oai-gray-300}{light gray} indicates the best result among open-source models. \textsuperscript{\dag} indicates results on \benchtiny set. \textbf{Right:} Results including the top-3 open-source models.}
    \label{tab:main_table}
    \vspace{-0.4cm}
\end{figure*}

\vspace{0.05cm}
\noindent\textbf{Human-in-the-loop Quality Review.} Despite human-annotated data sources and a meticulously designed QA generation methodology, certain ambiguities and errors inevitably persist, primarily due to inherent annotation errors in the source datasets. We implement a human-in-the-loop verification protocol spanning benchmark construction. This iterative quality assurance is bidirectional: when evaluators flag ambiguous or erroneous questions, we trace the error source and remove the problematic data sample or modify the meta-information, question template, or QA generation rule accordingly to rectify other erroneous questions stemming from the same source. Following each human review cycle, we update and iterate the benchmark until it satisfies our quality standards.

\vspace{-0.1cm}
\section{Evaluation on \bench}
\vspace{-0.1cm}
\label{sec:evaluation}

\subsection{Evaluation Setup}
\vspace{-0.1cm}
\noindent\textbf{Benchmark Models.}
We comprehensively evaluate 15 video-supporting MLLMs across diverse model families, encompassing various parameter scales and training recipes.
For proprietary models, we consider Gemini-1.5~\cite{team2024gemini} and GPT-4o~\cite{hurst2024gpto}.
For open-source models, we evaluate models from InternVL2~\cite{chen2024internvl2}, ViLA~\cite{lin2024vila}, LongViLA~\cite{xue2024longvila}, LongVA~\cite{zhang2024longva}, LLaVA-OneVision~\cite{li2024llavaov}, and LLaVA-Video~\cite{zhang2024video}. All evaluations are conducted under zero-shot settings and using each model's default prompts. To ensure reproducibility, we use greedy decoding for all models. 

\vspace{0.05cm}
\noindent\textbf{Metric Design.}
Based on whether the ground-truth answer is verbal or numerical, our tasks are suited to either a Multiple-Choice Answer (MCA) or Numerical Answer (NA) format (see \cref{fig:task_demonstration}). For MCA tasks, we follow standard practice~\cite{yue2024mmmu,fu2024video,hendrycks2021measuring} by using \textit{Accuracy} ($\mathcal{ACC}$), based on exact matching (with possible fuzzy matching), as the primary metric. For NA tasks, where models predict continuous values, accuracy via exact matching fails to capture the degree of proximity between model predictions and ground-truth answers. Therefore, we introduce a new metric, \textit{Mean Relative Accuracy} ($\mathcal{MRA}$), inspired by previous works~\cite{everingham2010pascal, lin2014microsoft, salton1986introduction}.
Specifically, for a NA question, given a model's prediction $\hat{y}$, ground truth $y$, and a confidence threshold $\theta$, relative accuracy is calculated by considering $\hat{y}$ correct if the relative error rate, defined as $|\hat{y}-y|/y$, is less than $1-\theta$.
As single-confidence-threshold accuracy only considers relative error in a narrow scope, $\mathcal{MRA}$ averages the relative accuracy 
across a range of confidence thresholds $\mathcal{C}=\{0.5, 0.55, \dots, 0.95\}$:
\vspace{-0.2cm}
\begin{equation}
\vspace{-0.2cm}
\begin{small}
    \mathcal{MRA} = \frac{1}{10}\sum_{\theta \in \mathcal{C}}\mathbbm{1} \left( \frac{|\hat{y} - y|}{y} < 1-\theta \right).
\end{small}
\end{equation}
$\mathcal{MRA}$ offers a more reliable and discriminative measurement for calculating the similarity between numerical predictions and ground truth values. 

\vspace{0.05cm}
\noindent\textbf{Chance Level Baselines.} We provide two baselines:
\begin{itemize}
    \item \textit{Chance Level (Random)} is the random selection accuracy for MCA tasks (and is inapplicable for NA tasks).
    \item \textit{Chance Level (Frequency)} represents the highest performance MLLMs would achieve by always selecting the most frequent answer for each task. This identifies performance gains that may result from inherently long-tailed answers or imbalanced multiple-choice distributions.
\end{itemize}

\vspace{0.05cm}
\noindent\textbf{Human Level Performance.}
We sample a subset of 400 questions (50 per task), which we will refer to as \benchtiny. Human evaluators independently answer each question, and their performance is evaluated using the above-mentioned metrics. For comparison, we also report \geminipro{}'s performance on \benchtiny. See  \cref{sec:supp_evaluation} for details on evaluation setups.

\begin{figure*}[t]
\centering
    \vspace{-0.5cm}
    \includegraphics[clip, trim=0cm 0cm 0cm 0cm, width=1\linewidth]{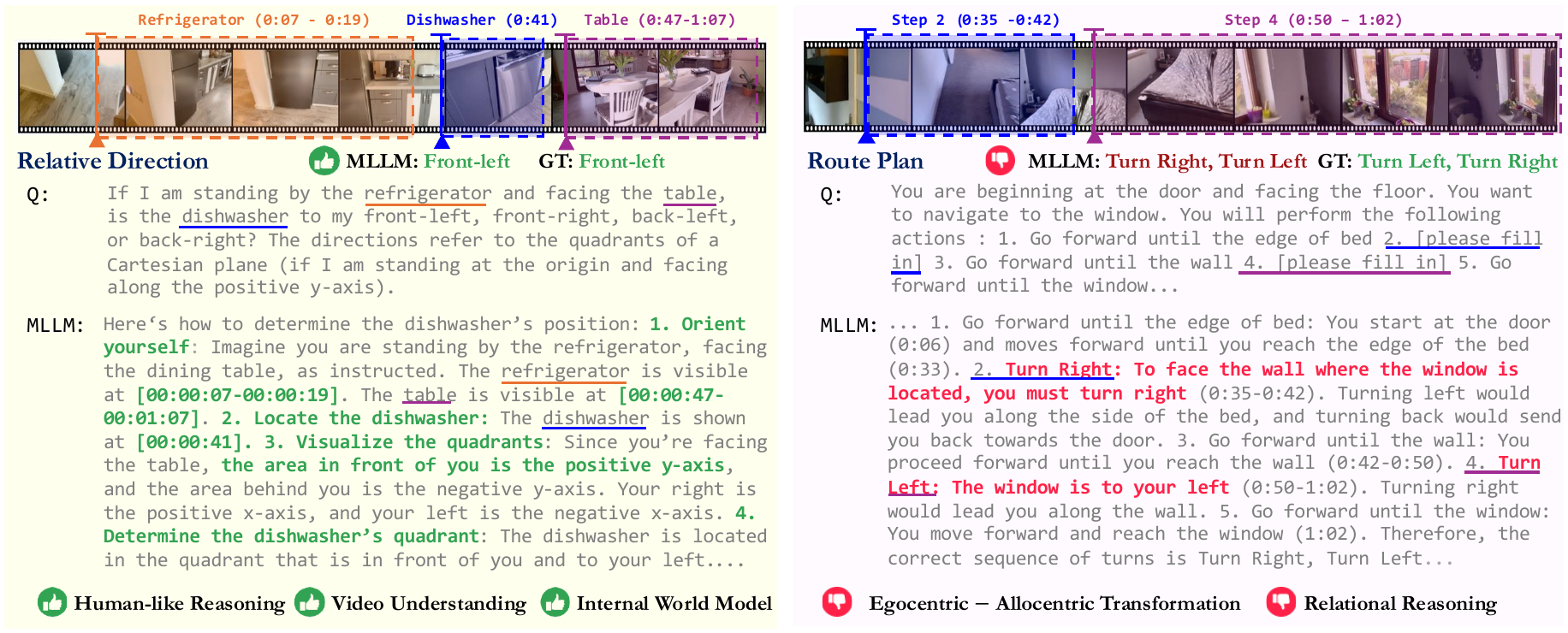}
    \vspace{-0.65cm}
    \caption{\textbf{Examples of how a MLLM thinks as seen in self-explanations}. While a MLLM exhibits strong video understanding and linguistic reasoning capabilities, its spatial reasoning capabilities are still developing.}
    \label{fig:linguistic_world_model_examples}
    \vspace{-4mm}
\end{figure*}

\subsection{Main Results}
\vspace{-0.1cm}
\cref{tab:main_table} shows overall model performance on \bench.
Our key observations are as follows:

\vspace{0.05cm}
\noindent\textbf{Human Level Performance.}
Not surprisingly, human evaluators achieve 79\% average accuracy on our benchmark, outperforming the best model by 33\%.
Notably, human performance on configuration and spatiotemporal tasks is remarkably high, ranging from 94\% to 100\%, indicating human intuitiveness. In contrast, the performance gap between humans and the best MLLM is much narrower on the three measurement tasks that require precise estimation of absolute distance or size, suggesting that MLLMs may have a relative strength in tasks requiring quantitative estimation. 

\vspace{0.05cm}
\noindent\textbf{Proprietary MLLMs.}
Despite a significant performance gap with humans, the leading proprietary model, Gemini-1.5 Pro, delivers competitive results. It surpasses the chance level baselines by a substantial margin and manages to approach human level performance in tasks such as absolute distance and room size estimation. It's worth noting that while human evaluators have years of experience in understanding the physical world spatially, MLLMs are only trained on 2D digital data like internet videos.

\vspace{0.05cm}
\noindent\textbf{Open-source MLLMs.}
Top-tier open-source models like LLaVA-Video-72B and LLaVA-OneVision-72B demonstrate highly competitive performance to closed-source models, trailing the leading Gemini-1.5 Pro by only 4\% to 5\%. However, the majority of open-source models (7$/$12) perform below the chance level baseline, indicating significant limitations in their visual-spatial intelligence.

\vspace{0.05cm}

\vspace{-0.1cm}
\section{How MLLMs \emph{Think in Space} Linguistically}
\vspace{-0.2cm}
\label{sec:think_linguistic}
To better understand when and why models succeed or fail and to elucidate the facets of visual-spatial intelligence they possess, we examine how MLLMs \textit{think in space} linguistically here and visually in \cref{sec:think_visual}. We begin by prompting the best-performing MLLM in \bench, \geminipro~\cite{team2024gemini}, to articulate its internal reasoning in language.

\subsection{Probing via Self-Explanations}
\vspace{-0.1cm}
\label{sec:linguistic_world_model}
Self-explanations are a prevailing approach on par with traditional model explanations like LIME saliency maps~\cite{ribeiro2016should} for understanding LLM-generated responses~\cite{huang2023can,lyu-etal-2023-faithful,gao-etal-2024-self} and are widely used in analyzing language model behavior~\cite{yue2024mmmu,parcalabescu2024measuring}. We randomly sample a subset of 163 incorrect answers, prompt the MLLM to provide explanations for the predicted answers, and carefully review them by hand.

\vspace{0.05cm}
\noindent\textbf{Case Studies.}
\cref{fig:linguistic_world_model_examples} presents self-explanations in both a success and an error case. In both examples, when thinking in space, the MLLM exhibits advanced video understanding, demonstrated by the impressive accuracy of its timestamped descriptions.
The model also forms correct step-by-step reasoning processes, outlining steps such as  ``orient yourself'', ``locate the dishwasher'' and ``visualize the quadrants'' for the relative direction task.
Furthermore, the construction of a global coordinate system (\cref{fig:linguistic_world_model_examples}, left) suggests that MLLMs may possess or build an implicit world model. Rather than using isolated frames, short clips, or random guesses, the MLLM used global spatial context and reasoning to infer correctly.

\begin{figure}[t]
    \centering
    % \vspace{-0.3cm}
    \includegraphics[width=1.05\linewidth]{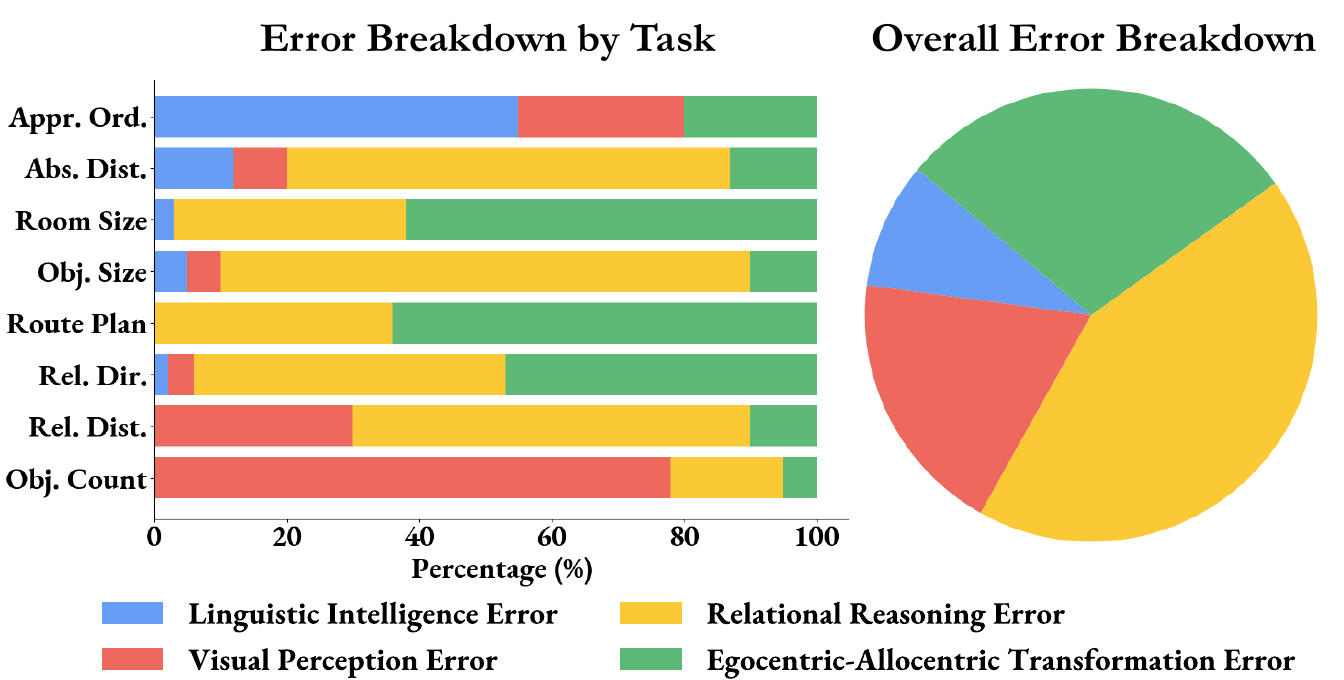}
    \vspace{-0.5cm}
    \caption{\textbf{Human-conducted analysis of errors by type.} Over 70\% of errors stem from faulty spatial reasoning capabilities.}
    \label{fig:error_analysis}
    \vspace{-0.4cm}
\end{figure}

In the incorrect example (\cref{fig:linguistic_world_model_examples}, right), we can identify faulty visual-spatial capabilities like egocentric-allocentric transformation and relational reasoning, as introduced in \cref{fig:vsi-taxonomy}. In the video, the camera pans right to shift the view from the edge of the bed to the wall and window. The model obeys this egocentric view, responding that ``to face the wall where the window is located, you must turn right'' instead of creating an allocentric view reflecting the reality that the route from door to bed means turning left.

\vspace{0.05cm}
\noindent\textbf{Error Analysis.}
To quantify and identify the main bottleneck for the best-performing MLLM on our benchmark, we analyze its errors on \benchtiny, categorizing them into four distinct types which arose from both our outlined visual-spatial capabilities (\cref{fig:vsi-taxonomy}) and a clear four-way categorization of errors upon examination:

\begin{enumerate}
    \item \textcolor{visual}{\textbf{Visual perception error}}, stemming from unrecognized objects or misclassified object categories;
    \item \textcolor{linguistic}{\textbf{Linguistic intelligence error}}, caused by logical, mathematical reasoning, or language understanding defects;
    \item \textcolor{relational}{\textbf{Relational reasoning error}} includes errors in spatial relationship reasoning, \ie, distance, direction, and size;
    \item \textcolor{egocentric}{\textbf{Egocentric-allocentric transformation error}}, resulting from an incorrect allocentric spatial layout or improper perspective-taking.
\end{enumerate}
As shown in \cref{fig:error_analysis}, around 71\% of errors are attributed to spatial reasoning (as ontologically conceived in \cref{fig:vsi-taxonomy}), which suggests that:
\infobox{Spatial reasoning is the primary bottleneck for MLLM performance on \bench.}
\vspace{0.1cm}

\noindent Further analysis and case studies are in \cref{sec:supp_error_analysis}. 

\subsection{Limits of CoT Methods in Visuospatial Tasks}
\label{sec:cot_limits_visuospatial}
Prompting techniques improve the reasoning and problem-solving abilities for large models across diverse tasks~\cite{jimenez2024swebench,wang2023voyager,huang2022language,suris2023vipergpt}. Their successes motivate us to investigate whether these linguistic prompting methods could also improve the visual-spatial capabilities of MLLMs in \bench. We investigate three prevailing prompting techniques (see \cref{sec:supp_cot_details} for more details):

\begin{itemize}
    \item \textit{Zero-Shot Chain-of-Thought (CoT).} Following~\cite{wei2022chain, kojima2022large}, we add ``Let's think step by step'' to the prompts.
    \item \textit{Self-Consistency w/ CoT.} We follow~\cite{wang2022self} and set the MLLM's temperature to 1.0 to encourage diverse reasoning and then take the majority consensus from five runs (employed w/ Zero-Shot CoT) as the final prediction.
    \item \textit{Tree-of-Thoughts (ToT).} Following the ``Creative Writting'' practice in~\cite{yao2024tree}, we divide reasoning into plan generation and answer prediction. The MLLM first drafts and selects a plan, then generates three candidate answers and selects the most confident one as prediction.
\end{itemize}

\begin{figure}[t]
    \centering
    \vspace{-0.3cm}
    \includegraphics[width=1\linewidth]{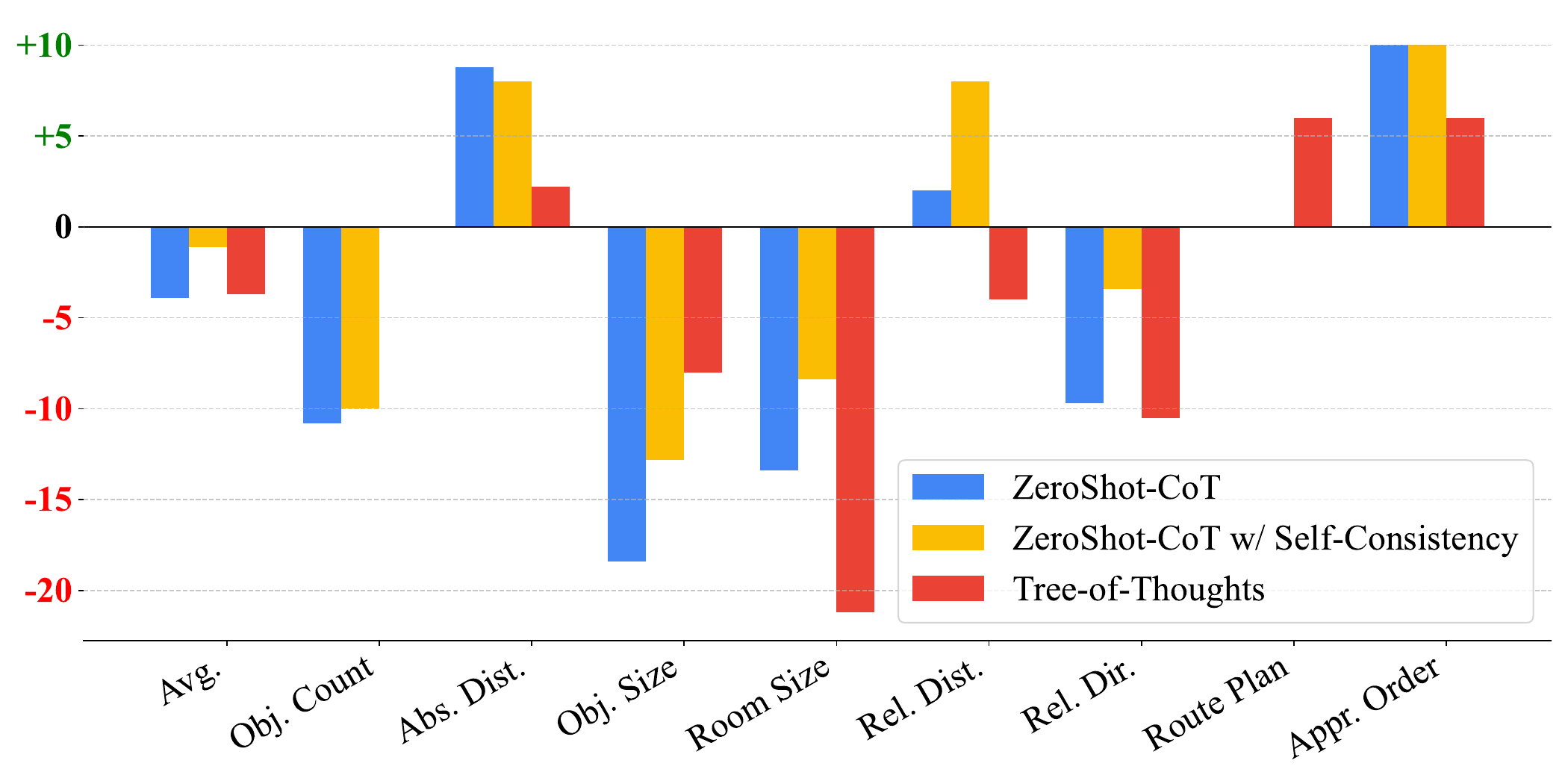}
    \vspace{-0.6cm}
    \caption{Relative improvements of \emph{CoT}, \emph{self-consistency} and \emph{Tree-of-Thought} compared to the baseline. All three prevailing prompting techniques fail on average on our benchmark, and, in some cases, task performance becomes \emph{much worse} after applying them.
    This implies that \bench cannot be solved by solely improving linguistic capabilities.} 
    \label{fig:chain_of_thought_prompting}
    \vspace{-0.2cm}
\end{figure}

As shown in \cref{fig:chain_of_thought_prompting}, surprisingly, all three linguistic reasoning techniques lead to performance degradation on \bench. Zero-Shot CoT and ToT reduce average performance by about 4\%, and self-consistency, though slightly better, still falls 1.1\% below the no-prompting baseline. The unilateral improvement in the appearance order and absolute distance estimation tasks is easily explained by their significant percentage of linguistic intelligence errors (see \cref{fig:error_analysis}). In contrast, the room size and object size tasks suffer a large 8\% to 21\% decrease, showing that encouraging a model to think more can be not just unreliable but downright harmful. Meanwhile, as shown in \cref{tab:videomme_cot}, ZeroShot CoT achieves a 1.6\% improvement on the general video understanding benchmark VideoMME~\cite{fu2024video}. Therefore, our results suggest that:
\infobox{Linguistic prompting techniques, although effective in language reasoning and general visual tasks, are harmful for spatial reasoning.
} 

\begin{table}[t]
    \vspace{0.01cm}
    \small
    \centering
    \begin{tabular}{lc}
        Case & Performance \\
        \toprule
        Gemini-1.5 Pro (w/o CoT) & 77.2 \\
        Gemini-1.5 Pro (w/~~ CoT) & 79.8 \\
    \end{tabular}
    \vspace{-0.2cm}
    \caption{\textbf{Gemini-1.5 Pro CoT performance on a 500-questions subset in VideoMME.} 
    }
    \label{tab:videomme_cot}
    \vspace{-0.55cm}
\end{table}

\vspace{-0.1cm}
\section{How MLLMs \emph{Think in Space} Visually}
\vspace{-0.1cm}
\label{sec:think_visual}
Since humans subconsciously build mental representations~\cite{tolman1948cognitive, nadel2008hippocampus} of space when reasoning spatially, we explore how MLLMs remember spaces.

\subsection{Probing via Cognitive Maps}
\label{sec:probe_cognitive_map}
\begin{figure}[t]
    \centering
    \vspace{-0.4cm}
    \includegraphics[width=1\linewidth]{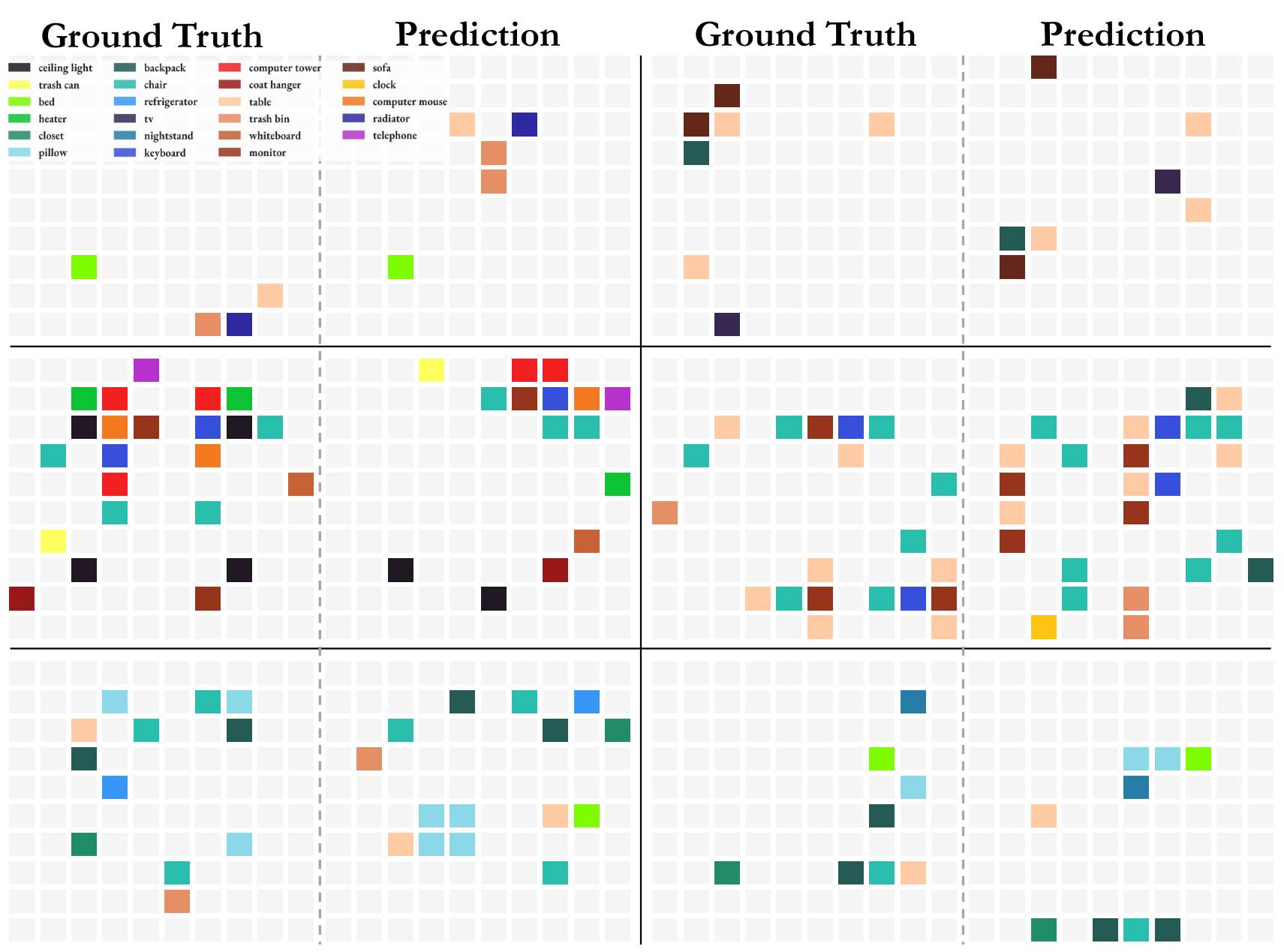}
    \vspace{-0.7cm}
    \caption{Visualization of cognitive maps from MLLM and GT. }    \label{fig:spatial_memory_analysis}
    \vspace{-0.4cm}
\end{figure}

\begin{figure}[t]
    \centering
    \includegraphics[width=1\linewidth]{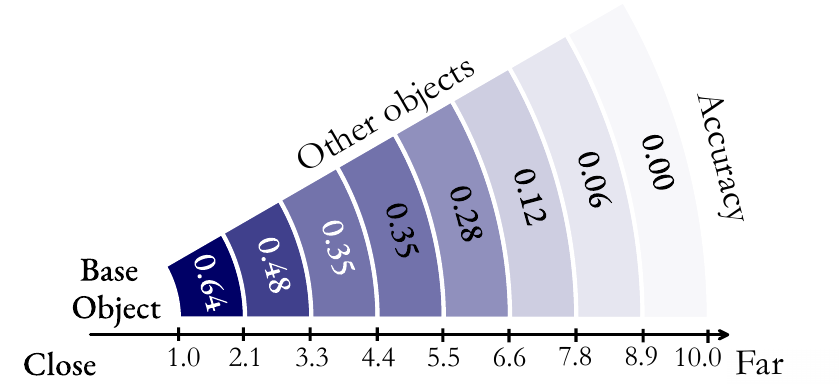}
    \vspace{-0.7cm}
    \caption{\textbf{Locality of the MLLM's predicted cognitive maps.} The MLLM's map-distance accuracy decreases dramatically with increasing object distance.}
    \label{fig:cog_map_distance}
    \vspace{-0.5cm}
\end{figure}

We prompt MLLMs to express their internal representations of the spaces they see using cognitive maps, a well-established framework for remembering objects in a set environment~\cite{Newcombe2024Spatial,tolman1948cognitive}. 
We prompt the best-performing MLLM, \geminipro, to predict object center positions within a 10 $\times$ 10 grid based on video input (see \cref{fig:cog_map_distance}\textcolor{linkblue}{b} for grid size ablation and \cref{sec:suppl_cog_map} for prompt). We present examples of the resulting cognitive maps in \cref{fig:spatial_memory_analysis}. 

To quantitatively assess these cognitive maps, we evaluate the Euclidean distance between all pairs of objects within each map. We consider the distance (on the grid) between two objects to be correct if it deviates by no more than one grid unit from the distance in the ground truth cognitive map. As shown in \cref{fig:cog_map_distance}, we divide the map-distances into eight distinct bins for analysis. Interestingly, we find that the MLLM achieves a remarkable 64\% accuracy in positioning adjacent objects within its cognitive map, indicating robust \emph{local} spatial awareness. However, this accuracy significantly deteriorates as the distance between two objects increases, which suggests that: 
\infobox{When remembering spaces, a MLLM forms a series of local world models in its mind from a given video, rather than a unified global model.}

\vspace{0.1cm}
\noindent This observation aligns with the challenge of forming a global space representation from discrete video frames, which is inherently difficult for MLLMs. While this task may not be trivial for humans either, it is likely that they can build such global space representations more accurately. 

\subsection{Better Distance Reasoning via Cognitive Maps}
\label{sec:reason_with_cognitive_map}
\vspace{-0.1cm}
Given the local awareness of MLLMs in remembering spaces (see \cref{fig:spatial_memory_analysis} and \cref{fig:cog_map_distance}) and the importance of mental imagery to how humans think in space, we investigate whether generating and using cognitive maps can help MLLMs' spatial reasoning in terms of \bench's relative distance task. This tests if the local distance awareness emerged through cognitive maps transfers to improved distance recall and reasoning. 

\begin{table}[ht]
    \vspace{-0.2cm}
    \small
    \centering
    \begin{subtable}[t]{.49\linewidth}
        \centering
        \vspace{-0.83cm}
        \setlength\tabcolsep{2pt}
        \scalebox{0.75}{
        \begin{tabular}{lc}
            Case & Rel. Dist Acc. \\
            \toprule
            w/o Cog. map & 46.0 \\
            w/ ~~Cog. map & 56.0 \\
            w/ ~~Cog. map (GT) & 66.0 \\
        \end{tabular}}
        \vspace{0.175cm}
        \caption{Cognitive map prompting.}
    \end{subtable}%
    \hfill
    \begin{subtable}[t]{.49\linewidth}
        \centering
        \setlength\tabcolsep{2pt}
        \scalebox{0.75}{
        \begin{tabular}{lcc}
            Cog. Map Src. & Size & Rel. Dist Acc. \\
            \toprule
            MLLM & 10 $\times$ 10 & 56.0 \\
            MLLM & 20 $\times$ 20 & 54.0 \\
            GT & 10 $\times$ 10 & 66.0 \\
            GT & 20 $\times$ 20 & 78.0 \\
        \end{tabular}}
        \vspace{-0.1cm}
        \caption{Cognitive map canvas size.}
    \end{subtable}
    \vspace{-0.3cm}
    \caption{\textbf{Relative distance task with cognitive map.} 
    }
    \label{tab:cog_map_prompting}
    \vspace{-0.3cm}
\end{table}

We prompt Gemini-1.5 Pro to first generate a cognitive map based on the given video and question, and then to use the predicted map to answer the question.
As shown in \cref{tab:cog_map_prompting}\textcolor{linkblue}{a}, we find that using mental imagery improves a MLLM's relative distance accuracy by 10\%. The 20\% to 32\% gain over baseline with the ground truth cognitive map underscores the importance of building accurate mental maps of a scene, which enforce a globally consistent topology, but indicates that such mental imagery is only one part of the puzzle, albeit a crucial one. These results point to building a mental spatial world model or cognitive map as a valuable pretext task or a promising solution for MLLMs to tackle visual-spatial reasoning.

\vspace{-0.1cm}
\section{Related Works}
\vspace{-0.1cm}
Apart from visual-spatial intelligence in \cref{sec:vsi}, we further ground our work in the following two related areas:

\vspace{0.1cm}
\noindent\textbf{MLLMs with Visual-Spatial Awareness}.
Building on the powerful language and reasoning abilities of LLMs~\cite{radford2018improving,radford2019language,brown2020language,touvron2023llama,touvron2023llama2,bai2023qwen,team2023gemini} and the feature extraction abilities of modern vision encoders~\cite{he2022masked,radford2021learning,oquab2024dinov}, MLLMs, especially visual MLLMs, exhibit unprecedented visual understanding capabilities~\cite{hurst2024gpto,team2024gemini,li2024llavaov,wang2024qwen2vl,zhang2024llavanextvideo,xue2024longvila, zhang2024redundancyrelevanceinformationflow}, a promising direction toward developing world models~\cite{liu2024world} and embodied agents~\cite{driess2023palm,mu2024embodiedgpt,kim2024openvla,cho2023spatially}.
However, grounding MLLMs in the real world presents significant challenges for models' visual-spatial intelligence, motivating recent efforts~\cite{yang2024virl,chen2024spatialvlm,cheng2024spatialrgpt,cai2024spatialbot,liu2024coarsecorrespondenceelicit3d,li2024topviewrs,zhu2024llava3d,han2024videoespresso}.
Unlike prior works, which primarily focus on understanding spatial information through 2D images~\cite{tang2024sparkle,ramakrishnan2024does,yang2023setofmarkpromptingunleashesextraordinary} or solely language~\cite{wu2024visualization,rozanova2021grounding,yamada2024evaluating,momennejad2024evaluating,wu2024visualization}, our work assesses models' visual spatial intelligence using real-world videos, which more closely mirrors human understanding of the world and application scenarios for embodied agents.

\vspace{0.1cm}
\noindent\textbf{Benchmarking MLLMs on Video}.
With MLLMs displaying impressive performance on still-images across perception, reasoning, and multi-disciplinary tasks~\cite{yue2024mmmu,liu2025mmbench,li2024seed,yu2023mmvet}, there is increasing interest in evaluating MLLMs' video understanding capabilities~\cite{fu2024video,mangalam2023egoschema,li2024mvbench,ning2023videobench,liu-etal-2024-tempcompass,li2023vitatecs,fang2024mmbenchvideo,ye2024mmegobuildingegocentricmultimodal,majumdar2024openeqa,linghu2024multi,wang2024picture}.
For example, Video-MME~\cite{fu2024video} comprehensively evaluates MLLMs across various video-related tasks, including recognition and perception.
EgoSchema~\cite{mangalam2023egoschema} and OpenEQA~\cite{o2023open} evaluate MLLMs' understanding abilities using egocentric videos.
Despite their significance, most prior works focus on content-level understanding~\cite{fu2024video,li2024mvbench,mangalam2023egoschema,ning2023videobench}, which primarily serves as a temporal extension of 2D image understanding without 3D spatial consideration. 
% Extending beyond prior benchmarks, our work establishes a testbed evaluating models' 3D video-based visual-spatial intelligence,  using video as an interface to understand the real world.
Extending beyond prior benchmarks, our work focuses on spatial intelligence and requires core spatial capabilities like visual working memory and implicit scene reconstruction.

\vspace{-0.1cm}
\section{Discussion and Future Work}
\vspace{-0.2cm}
We study how models see, remember, and recall spaces by building \bench  
and investigating the performance and behavior of MLLMs on it. Our analysis of how MLLMs think in space linguistically and visually identifies existing strengths (\eg, prominent perceptual, temporal, and linguistic abilities) and bottlenecks for visual-spatial intelligence (\eg, egocentric-allocentric transformation and relational reasoning). While prevailing linguistic prompting methods fail to improve spatial reasoning, building explicit cognitive maps does enhance the spatial distance reasoning of MLLMs.  
Future avenues of improvement include task-specific fine-tuning, developing self-supervised learning objectives for spatial reasoning, or visuospatial-tailored prompting techniques for MLLMs.

\vspace{0.1cm}
\noindent\textbf{Acknowledgments.}
We thank Ellis Brown, Ryan Inkook Chun, Youming Deng, Oscar Michel, Srivats Poddar, Xichen Pan, Austin Wang, Gavin Yang, and Boyang Zheng for their contributions as human annotators and evaluators. We also thank Fred Lu for proofreading our manuscript. 
We also thank Chen Feng, Richard Tucker, Noah Snavely, Leo Guibas, and Rob Fergus for their helpful discussions and feedback. This work was mainly supported by the Open Path AI Foundation, Google TPU Research Cloud (TRC) program, and the Google Cloud Research Credits program (GCP19980904). SX also acknowledges support from Intel AI SRS, IITP grant funded by the Korean Government (MSIT) (No. RS-2024-00457882, National AI Research Lab Project), Amazon Research Award, and NSF Award IIS-2443404.

{
    \small
    \bibliographystyle{ieeenat_fullname}
    \bibliography{main}
}

\clearpage
\appendix

\section{Appendix Outline}
\label{sec:appendix}
In these supplementary materials, we provide:
\begin{itemize}
\item Technical details about \bench construction and our linguistic and visual analysis (\cref{sec:supp_implementation_details}); 
\item Evaluation setup and full evaluation results for \bench sub-experiments (\cref{sec:supp_evaluation});
\item Analysis on input sequencing and repetition (\cref{sec:supp_analysis_exp});
\item Additional visualization results (\cref{sec:supp_visualization_results}).
\end{itemize}

\section{Technical Details for \bench Construction and Analysis}
\label{sec:supp_implementation_details}
In this section, we provide more technical details on the construction of \bench and analyzing MLLM thinking via self-explanations, Chain-of-Thought-based methods, and cognitive maps.

\subsection{\textbf{\bench} Construction Pipeline}
\label{sec:supp_bench_construct}
Here, we discuss the concrete setup for each stage in the benchmark construction pipeline.

\vspace{0.1cm}
\noindent\textbf{Dataset Collection and Unification.} We curate our evaluation dataset by collecting 150 samples from ARKitScenes~\cite{dehghan2021arkitscenes}, 50 samples from ScanNet++~\cite{yeshwanth2023scannet++}, and 88 samples from ScanNet~\cite{dai2017scannet}. For video processing, we convert ScanNet's individual frames into continuous videos at 24 FPS, while subsampling ScanNet++ and ARKitScenes videos to 30 FPS. All videos are standardized to a resolution of 640 $\times$ 480 pixels. Given that ARKitScenes contains videos with varying orientations, we normalize their rotation to maintain a consistent upward orientation across all samples.

Due to varying annotation structures across the three datasets, we unify them into a standardized meta-information format for each scene with the following attributes: \textit{dataset}, \textit{video path}, \textit{room size}, \textit{room center}, \textit{object counts}, and \textit{object bounding boxes}. The room size is calculated by the Alpha shape algorithm\footnote{\url{https://en.wikipedia.org/wiki/Alpha_shape}} with the scene's point cloud. The room center is calculated as the geometric center of the minimal bounding box of the scene's point cloud. Object counts record the number of instances for each category. As for the object bounding boxes, we unify different annotation formats to the format of \texttt{OrientedBoundingBox} in Open3D~\cite{Zhou2018open3d}.

For the categories included in the meta-information, we carefully curate a subset of categories from the three source datasets. Since our benchmark aims to evaluate the visual-spatial intelligence of MLLMs, we exclude both rare categories and those with extremely small object sizes to reduce perceptual challenges. Additionally, we implement category remapping to ensure vocabulary consistency and intuitive understanding across 
the benchmark. This category remapping is also iteratively refined during human review.

\begin{table*}[t]
    \renewcommand{\arraystretch}{1.5} 
    
    \begin{tabular}{r|p{13.5cm}}
        \textbf{Task} & \textbf{Question Template} \\
        \hline
        Object Counting & 
        \textit{How many \textcolor{red}{\{category\}}(s) are in this room?} \\
        \hline
        Relative Distance & 
        \textit{Measuring from the closest point of each object, which of these objects (\textcolor{red}{\{choice a\}}, \textcolor{red}{\{choice b\}}, \textcolor{red}{\{choice c\}}, \textcolor{red}{\{choice d\}}) is the closest to the \textcolor{red}{\{category\}}?} \\
        \hline
        Relative Direction & 
        To create a comprehensive test of relative direction, three difficulty levels were created:
        \begin{itemize}
            \item \textbf{Easy:} \textit{If I am standing by the \textcolor{red}{\{positioning object\}} and facing the \textcolor{red}{\{orienting object\}}, is the \textcolor{red}{\{querying object\}} to the left or the right of the \textcolor{red}{\{orienting object\}}?}
            \item \textbf{Medium:} \textit{If I am standing by the \textcolor{red}{\{positioning object\}} and facing the \textcolor{red}{\{orienting object\}}, is the \textcolor{red}{\{querying object\}} to my left, right, or back? An object is to my back if I would have to turn at least 135 degrees in order to face it.}
            \item \textbf{Hard:} \textit{If I am standing by the \textcolor{red}{\{positioning object\}} and facing the \textcolor{red}{\{orienting object\}}, is the \textcolor{red}{\{querying object\}} to my front-left, front-right, back-left, or back-right? Directions refer to the quadrants of a Cartesian plane (assuming I am at the origin and facing the positive y-axis).} 
            \vspace{-0.4cm}
            \end{itemize} \\
        \hline
        Appearance Order & 
        \textit{What will be the first-time appearance order of the following categories in the video: \textcolor{red}{\{choice a\}}, \textcolor{red}{\{choice b\}}, \textcolor{red}{\{choice c\}}, \textcolor{red}{\{choice d\}}?}
         \\
        \hline
        Object Size & 
        \textit{What is the length of the longest dimension (length, width, or height) of the \textcolor{red}{\{category\}}, measured in centimeters?}
         \\
        \hline
        Absolute Distance & 
        \textit{Measuring from the closest point of each object, what is the direct distance between the \textcolor{red}{\{object 1\}} and the \textcolor{red}{\{object 2\}} (in meters)?}
         \\
        \hline
        Room Size & 
        \textit{What is the size of this room (in square meters)? 
        If multiple rooms are shown, estimate the size of the combined space.}
         \\
        \hline
        Route Plan & 
        \textit{You are a robot beginning at \textcolor{red}{\{the bed facing the tv\}}. You want to navigate to \textcolor{red}{\{the toilet\}}. You will perform the following actions (Note: for each [please fill in], choose either `turn back,' `turn left,' or `turn right.'): \textcolor{red}{\{1. Go forward until the TV 2. [please fill in] 3. Go forward until the shower 4. [please fill in] 5. Go forward until the toilet.\}} You have reached the final destination.} 
         \\
    \end{tabular}
    \vspace{-.3cm}
    \caption{\textbf{Question Templates for tasks in \bench.} We replace the \textcolor{red}{highlighted} part in the question template from scene to scene to construct our benchmark. Note that a complete example question is provided for Route Plan.}
    \vspace{-.5cm}
    \label{tab:tasks}
\end{table*}

\vspace{0.1cm}
\noindent\textbf{QA-Pair Generation.} Each QA-pair contains the following attributes: \textit{question ID}, \textit{source dataset}, \textit{task type}, \textit{video path}, \textit{question}, \textit{multiple-choice options w/ letter answer}, and \textit{verbal or numerical ground truth}. Of the eight tasks in \bench, the QA-pairs for seven tasks are derived from the unified meta-information and the Route Plan QA-pairs from human-annotated routes. 

We evaluate the multiple-choice answer (MCA) tasks via accuracy and the numerical-answer (NA) tasks via mean relative accuracy ($\mathcal{MRA}$), but our VQA dataset also includes generated multiple-choice options and letter answers for the NA tasks. The generated multiple-choice options are sampled between a lower and upper bound factor of the ground truth numerical answer and are re-sampled if any two options are within a given threshold of each other. We sub-sample the number of questions for each scene for each task to prevent over-representation of any scene or task and to create a more balanced dataset. For MCA tasks, the letter answers are distributed as uniformly as possible. 

For the \textit{object counting} task, objects with counts of one are not included. For the \textit{relative distance} task, only unique-instance objects are used for the primary category; multiple-instance objects are allowed for the object choices. If there are multiple instances of an object category, the minimum absolute distance to the primary object is used. If any of the four option distances are within a threshold (30 cm for rooms with size greater than 40 sq m, 15 cm otherwise) of each other, the question is considered ambiguous. For the \textit{relative direction} task, to make sure the direction is clear, questions are considered ambiguous if they violate lower and upper bounds on the distance between any two objects or a threshold for proximity to angle boundaries. For the \textit{appearance order} task, first appearance is considered to be the timestamp where the number of object pixels cross a set threshold, and timestamps too close together are considered ambiguous. For the \textit{object size} task, the ground truth is taken as the longest dimension of the unique object's bounding box. For the \textit{room size} task, room size is calculated by the alpha shape algorithm, as specified earlier. For the \textit{absolute distance} task, we first uniformly sample points within the bounding boxes of the two objects. The distance is the minimum Euclidean distance among pairwise points. For the \textit{route planning} task, humans construct routes given a template and instructions to choose any two unique objects as the start and end position, respectively, such that the route between them can be described in approximately two to five movements. Routes are comprised of two actions: ``Go forward until [unique object]'' and ``Turn [left / right / back]''. After collection, filtering and standardization are done. In the question, the "turn" directions are replaced with ``[please fill in]''.

The question templates for the generation of each task are listed in \cref{tab:tasks}.

\vspace{0.1cm}
\noindent\textbf{Human-in-the-loop Quality Review.} The quality review process occurs throughout two stages of our pipeline. During dataset collection, we manually filter the validation set by removing scenes with a high ratio of incomplete 3D mesh reconstruction that could misalign 3D annotations with visible video content. After generating scene meta-information, we manually verify its correctness, with a specific focus on ensuring the correctness of \textit{object counts}.

In the QA pairs generation stage, we customize a web interface for human quality review. Human evaluators are asked to answer the benchmark questions without prior knowledge of the correct answers. They flag QA pairs where they believe the answers are incorrect. When evaluators identify ambiguous or erroneous questions, we trace the source of the errors and take corrective actions, such as removing problematic data samples or adjusting the meta-information, question templates, or modifying QA generation rules to prevent similar issues in the future. We iterate this procedure multiple times to ensure the quality.

\subsection{Probing MLLM via Self-Explanations}
Here, we provide more concrete implementations for the self-explanations and error analysis. 

\vspace{0.1cm}
\noindent\textbf{Self-Explanations.} To conduct error analysis on a model's reasoning chains behind its predictions, we explicitly extract the reasoning chains that support the model's question-
answering process.
Specifically, after the model predicts an answer to a given question, it is further prompted with \textit{``Please explain your answer step by step.''} to generate the internal rationale leading to its prediction. It is important to note that this process is fundamentally different from \textit{Chain-of-Thought} reasoning, where the model is asked to generate reasoning chains first and then predict the answer.

\vspace{0.1cm}
\noindent\textbf{Error Analysis.} For error analysis, we manually review within \benchtiny all error cases for tasks in multiple-choice answers and the bottom half of the worst-performing cases for tasks in numerical answers, which totals 163 samples.
For each error case, human examiners are required to classify its primary error into one of four primary categories: \textit{visual perception error}, \textit{linguistic intelligence error}, \textit{relational reasoning error}, and \textit{egocentric-allocentric transformation error}. If an incorrect prediction is attributed to multiple reasons, it is proportionally assigned as 
$\frac{1}{n}$ to each applicable category, where $n$ is the number of error categories.

\subsection{Implementation Details of CoT Methods}
\label{sec:supp_cot_details}
As detailed in our paper, we evaluate several advanced linguistic prompting methods on our benchmark, including \textit{Chain-of-Thought}, \textit{Self-Consistency}, and \textit{Tree-of-Thoughts}. In this section, we elaborate on the implementation details of these three methods.

\begin{itemize}
    \item \textit{Chain-of-Thought} prompting. Following Zero-shot-CoT~\cite{wei2022chain,kojima2022large}, we append the phrase ``Let's think step by step.'' to each question to elicit step-by-step reasoning from the large language model. The temperature, top-p, and top-k parameters are set to 0, 1, and 1, respectively. After the model generates its prediction, we initiate an additional turn of dialogue to prompt the model to extract its answer explicitly (\eg, the letter corresponding to the correct option for multiple-choice questions or a numerical value for numerical questions). This approach mitigates errors arising from fuzzy matching.
    \item \textit{Self-Consistency w/ CoT}. In line with Self-Consistency~\cite{wang2022self}, we prompt MLLMs to generate multiple answers for a given question under Zero-shot-CoT~\cite{kojima2022large} prompting. To encourage diversity among runs, we set the temperature to 0.7, top-p to 1, and top-k to 40. Initially, the model is prompted to provide an answer with step-by-step reasoning (using Zero-shot-CoT). As with Zero-shot-CoT, an additional dialogue turn is added to explicitly extract the prediction from the model's response. For each question, we perform 5 independent runs and take the majority prediction as the final answer.
    \item \textit{Tree-of-Thoughts}. Inspired by the ``Creative Writing'' practice in \cite{yao2024tree}, we divide the problem-solving process into two steps: plan generation and answer prediction. The temperature, top-p, and top-k parameters remain consistent with the Self-Consistency setup. For the plan generation step, we ask the model to generate 3 distinct plans to answer the given question. We then start a new dialogue and prompt the model to select the most promising plan based on the video, the question and the generated plans. This voting process is repeated 3 times, with the majority-selected plan chosen for the next step. In the answer prediction step, based on the video and the selected plan, the model is asked to predict the answer. Similar to the previous step, 3 independent predictions are generated, and the model votes 3 times to determine the most confident answer. A majority vote determines the final prediction.
\end{itemize}
\cref{fig:supp_zero_shot_cot}. \cref{fig:supp_zero_shot_self-consistency}, and \cref{fig:supp-tot} illustrate these three prompting techniques and model outputs under the different strategies.

\subsection{Cognitive Map}
\label{sec:suppl_cog_map}
\noindent\textbf{Generation.} To generate the cognitive map for each video, we specify the target categories of interest and prompt the MLLM to predict the central position for each of these categories. The following prompt is used:

\begin{tcolorbox}[colback=black!5!white,colframe=black!75!black,title=Cognitive Map Prompt]

[Task]\\
This video captures an indoor scene. Your objective is to identify specific objects within the video, understand the spatial arrangement of the scene, and estimate the center point of each object, assuming the entire scene is represented by a 10x10 grid.\\

[Rule]\\
1. We provide the categories to care about in this scene: \{categories\_of\_interest\}. Focus ONLY on these categories. \\
2. Estimate the center location of each instance within the provided categories, assuming the entire scene is represented by a 10x10 grid. \\
3. If a category contains multiple instances, include all of them.\\
4. Each object's estimated location should accurately reflect its real position in the scene, preserving the relative spatial relationships among all objects.\\

[Output]\\
Present the estimated center locations for each object as a list within a dictionary.
STRICTLY follow this JSON format:
\{"category name": [(x\_1, y\_1), ...], ...\}
\end{tcolorbox}

For the categories of interest, we include all potential categories as shown in \cref{fig:spatial_memory_analysis} and \cref{fig:cog_map_distance}. Such setup facilitates our focus on assessing the spatial awareness of the MLLM rather than its perceptual capabilities. 
In contrast, for benchmark tasks such as evaluating relative distance (as shown in \cref{tab:cog_map_prompting}), we restrict the provided categories to those explicitly mentioned in each question. This ensures that no additional information apart from the question is included.

\vspace{0.2cm}
\noindent\textbf{Distance Locality Calculation.} To quantitatively evaluate the cognitive maps, we measure inter-category distances as illustrated in \cref{fig:cog_map_distance}. Specifically, for each category, we compute its Euclidean distance to all other categories. When a category contains multiple objects, we define the inter-category distance as the shortest distance between any two objects from the respective categories. We perform these distance calculations on both MLLM-predicted and ground truth cognitive maps and consider an MLLM's predicted distance between two categories to be correct if it differs from the ground truth distance by no more than one grid unit. We apply this evaluation process across all cognitive maps and group the distances into eight bins to calculate the average accuracy on different bins.

\subsection{Cognitive Map on More MLLMs}
We evaluate two more MLLMs, LLaVA-Video-7B and LLaVA-Video-72B. \cref{tab:cognative_map_exps} validates our Sec.~\ref{sec:probe_cognitive_map} finding of significantly stronger local than global accuracy. Regarding Sec.~\ref{sec:reason_with_cognitive_map}, as shown in \cref{tab:cog_map_rel_dist}, LLaVA-Video-72B achieves an 8\% performance gain. In contrast, LLaVA-Video-7B performance decreases, likely due to its limited model capacity, which impairs cog. map prediction (\cref{tab:cognative_map_exps} shows its suboptimal acc. on cog. map compared to Gemini-1.5 Pro and LLaVA-Video-72B).

\begin{table}[h]
    % \small
    \centering
    \vspace{0.05cm}
    \setlength\tabcolsep{2pt}
    \scalebox{0.57}{
    \begin{tabular}{l|cccccccc}
    % \hline
    Distance & [1.0, 2.1] & (2.1, 3.3] & (3.3, 4.4] & (4.4, 5.5] & (5.5, 6.6] & (6.6, 7.8] & (7.8, 8.9] & (8.9, 10.0] \\
    \hline
    Gemini-1.5 Pro & 0.64 & 0.48 & 0.35 & 0.35 & 0.28 & 0.12 & 0.06 & 0.00 \\
    LLaVA-Video-72B & 0.59 & 0.45 & 0.42 & 0.30 & 0.15 & 0.23 & 0.16 & 0.00 \\
    LLaVA-Video-7B & 0.50 & 0.43 & 0.34 & 0.29 & 0.19 & 0.18 & 0.14 & 0.00 \\
    % \hline
    \end{tabular}}
    \vspace{-0.1cm}
    \caption{\textbf{Locality of cognitive maps.}}
    \vspace{-0.3cm}
    \label{tab:cognative_map_exps}
\end{table}

\begin{table}[h!]
    \small
    \vspace{-0.2cm}
    \centering
    \setlength\tabcolsep{7pt}
    \renewcommand{\arraystretch}{1.0}
    \begin{tabular}{l|cc}
        Models & LLaVA-Video-72B & LLaVA-Video-7B \\
    \hline
    \emph{w/o.} Cog. Map & 36.0 & 40.0 \\
    \emph{w/.}~~ Cog. Map & 42.0  & 32.0 \\
    \end{tabular}
    \vspace{-0.1cm}
    \caption{\textbf{Rel. dist. task with cognitive maps.}}
    \vspace{-0.3cm}
    \label{tab:cog_map_rel_dist}
\end{table}

\section{Evaluation Details}
\label{sec:supp_evaluation}

\subsection{General Evaluation Setup}

Our evaluation processes are primarily conducted using the \texttt{LMMs-Eval} project~\cite{zhang2024lmms}.
To ensure reproducibility, unless otherwise specified, we adopt a greedy decoding strategy for all models (\emph{i.e.}, the temperature is set to 0, and both top-p and top-k are set to 1). The input for the models is formatted as follows: {\small\texttt{[Video Frames][Pre-prompt][Question][Post-prompt]}}, where {\small\texttt{Question}} includes the question and any available options. The specific {\small\texttt{Pre-prompt}} and {\small\texttt{Post-prompt}} for different models and question types are detailed in \cref{tab:evaluation_prompt}.

\subsection{Human Evaluation Setup}
During the evaluation of human-level performance on \benchtiny, human evaluators are allowed unlimited time to answer questions to the best of their ability. They receive both the questions and corresponding videos simultaneously and can review the videos multiple times to gather comprehensive information. We do not restrict the number of times evaluators can review videos for two key reasons. First, MLLMs auto-regressively generate answers, enabling them to analyze videos repeatedly during the response generation process. Second, MLLMs are designed to achieve and exceed typical human-level performance for practical real-world applications.

In addition, we provide the human evaluation on another \bench subset with 560 samples optimized to minimize the average performance gap between this subset and full set for all MLLMs. As shown in \cref{tab:560_subset_results}, this subset has an average performance discrepancy compared to full set (see \cref{tab:main_table}) just 0.5\% and a maximum of 2.9\%.

\subsection{Number of Frames Setup}

\begin{figure}[t]
    \centering
    \includegraphics[width=0.8\linewidth]{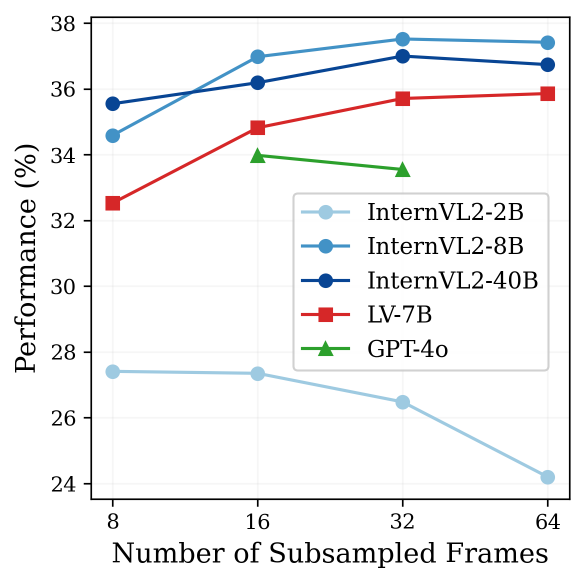}
    \vspace{-0.2cm}
    \caption{\textbf{Analysis of different \# sampled frames.}}
    \vspace{-0.5cm}
    \label{fig:supp-sampled-frames}
\end{figure}

Typically, MLLMs subsample a fixed number of frames for evaluation. For all open-source models and the GPT-4 API, following~\cite{zhang2024lmms}, we manually sample video frames from the entire video at evenly spaced time intervals. For the Gemini API, we follow its instructions, uploading and feeding the entire video to the model. The number of frames used for each model are provided in \cref{tab:number_of_frames_setup}. We believe that frame sampling strategies are a model design choice separate from the benchmark design.
Established benchmarks (\eg, VideoMME~\cite{fu2024video} and EgoSchema~\cite{mangalam2023egoschema}) also employ default sampling, reinforcing this perspective.
In addition, as shown in the \cref{fig:supp-sampled-frames}, the \# of sampled frames only marginally affects performance---it is not the primary bottleneck.

\subsection{More Evaluation Results}
Here, we provide more evaluation results on our benchmark, including blind evaluation results, the Socratic LLMs, the full evaluation results of \benchtiny, and vision-enabled $-$ vision-disabled results. 

\begin{figure}[t]
\centering
    \includegraphics[clip,trim=0cm 0cm 0cm 0cm,width=1.0\linewidth]{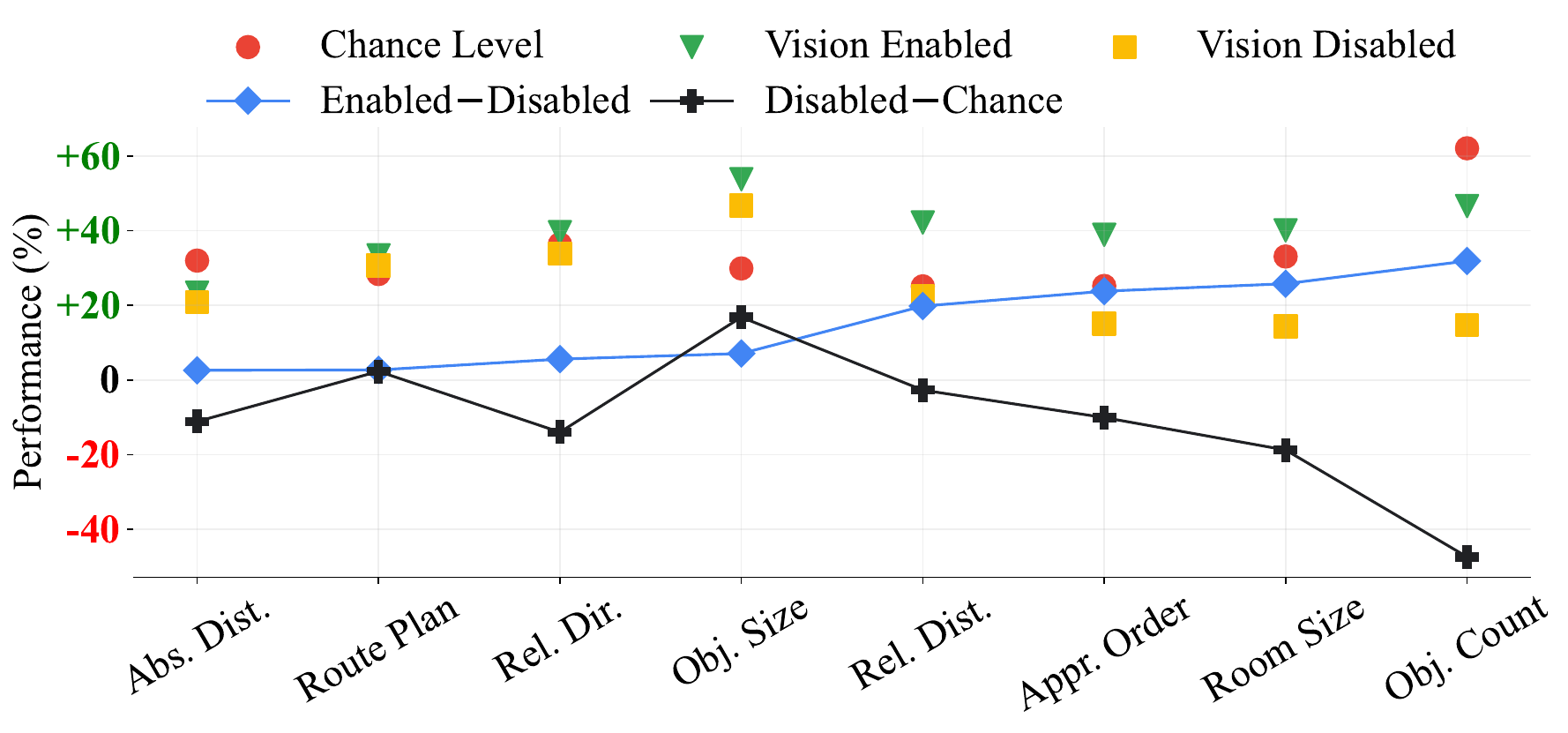}
    \vspace{-8mm}
     \caption{\textbf{Performance comparisons between Vision Enabled (w/ video), Vision Disabled (w/o video) and Chance Level (Freq.).} Enabled$-$Disabled indicates the gap between Vision Enabled and Vision Disabled, and Disabled$-$Chance betokens the gap between Vision Disabled and Chance Level (Freq.). Tasks are sorted by Enable$-$Disable for better understanding.} 
    \label{fig:video_gains}
    \vspace{-4mm}
\end{figure}

\vspace{0.1cm}
\noindent\textbf{Blind Evaluation.} 
We compare MLLMs' performance against  ``Chance Level (frequency)'' and ``Vision Disabled'' (blind) results, using averages across six of the strongest models (3 open-source and 3 closed-source).
As shown in \cref{fig:video_gains}, the consistent improvements in ``Enabled$-$Disabled'' 
and general degradation in ``Disabled$-$Chance'' demonstrates that video is essential and beneficial for our \bench, with blind models performing below chance level.
Meanwhile, MLLMs struggle to improve beyond chance level in the absolute distance estimation, route plan, and relative direction tasks, whether vision is enabled or not, underscoring the difficulty of these tasks.
Note that on object size, ``Vision Disabled'' models already significantly outperform chance level, likely due to common-sense knowledge learned during language model training.

In addition, as shown in \cref{tab:complete_blind_evaluation_results}, we present the evaluation results for all MLLMs on \bench. Generally, larger variants within the same model family often demonstrate better performance in blind evaluations, as seen in comparisons such as Gemini-1.5 Flash \vs Gemini-1.5 Pro and VILA-1.5-8B \vs VILA-1.5-40B. The blind evaluation also highlights LLM biases across tasks. For instance, LongVILA-8B achieves 47.5\% accuracy on the object count task, benefiting from a bias that frequently leads it to predict 2 as the answer.

\vspace{0.05cm}
\noindent\textbf{Socratic LLMs with Frame Captions.} Following OpenEQA~\cite{majumdar2024openeqa} and HourVideo~\cite{chandrasegaran2024hourvideo}, we implement a Socratic variant of GPT-4o using LLaVA-Video-72B as the captioner and GPT-4o as the answering LLM. As shown in \cref{tab:socratic_table}, Socratic lags behind the standard GPT-4o by 4.7\%.

\begin{table}[h]
    % \vspace{-0.3cm}
    \centering
    \setlength\tabcolsep{6pt} % Default value: 6pt
    \renewcommand{\arraystretch}{1.0} % Adjusts the row height
    \hspace*{-0.65cm}
    \scalebox{1.0}{
    \begin{tabular}{l|ccc}
    GPT-4o & Standard & Socratic & Blind  \\
    \hline
    Avg. & 34.0 & 29.3 & 14.5 \\
    \end{tabular}}
    \caption{\textbf{Socratic LLMs with Frame Captions.}}
    \label{tab:socratic_table}
    % \vspace{-0.5cm}
\end{table}

\vspace{0.05cm}
\noindent\textbf{\benchtiny Results.} As shown in \cref{tab:complete_subset_results}, we provide the evaluation results of all models on \benchtiny. The rankings and average accuracy of MLLMs on \benchtiny remain consistent to the results reported in \cref{tab:main_table}. This consistency suggests that the human evaluation and analysis results conducted on \benchtiny are reliable.

\vspace{0.1cm}
\noindent\textbf{Vision Enabled $-$ Vision Disabled.} \cref{tab:vision_enabled_disable} presents the improvement of MLLMs from using visual signals to answer \bench. Almost all MLLMs obtain improvements from visual signals, with notable improvements in tasks such as object count, room size, relative distance and appearance order. 

\section{Input Sequencing and Repetition Analysis}
\label{sec:supp_analysis_exp}
Human performance in visual problem-solving improves when they know the question before viewing the visual content, as it helps direct their attention to relevant visual cues. However, current MLLMs typically rely on a visual-first paradigm~\cite{team2024gemini,liu2024visual}, leading us to examine how the presentation order of video-question pairs impacts model performance. To investigate, we conduct experiments using Gemini-1.5 Pro on \benchtiny.

\vspace{0.1cm}
\noindent\textbf{MLLM's performance degrades with question-first paradigm.}
As shown in \cref{tab:video_text_ordering} (a), switching to a video-first approach results in a 2.5\% decrease in overall performance for Gemini compared to the question-first approach.

\begin{table}[t!]
    \vspace{-0.25em}
    \small
    \centering
    \begin{subtable}[t]{.45\linewidth}
        \centering
        
        \begin{tabular}{lc}
            Order & Avg. \\
            \hline
            Video first & \textbf{48.8} \\
            Question first & 46.3 \\
        \end{tabular}
        \caption{Input Sequence}
    \end{subtable}%
    \hfill
    \begin{subtable}[t]{.45\linewidth}
        \centering

        \begin{tabular}{lc}
            \# Times & Avg. \\
            \hline
            1 & 48.8 \\
            2 & \textbf{50.9} \\
        \end{tabular}
        \caption{Video Repetition Times}
    \end{subtable}
\vspace{-0.3cm}
\caption{Ablations on the video input sequence and repetition.}
\label{tab:video_text_ordering}
\vspace{-0.25em}
\end{table}

\vspace{0.1cm}
\noindent\textbf{MLLM benefits from multiple video views.}
In addition, humans often improve their VQA performance by reviewing visual content multiple times, inspiring us to implement a similar setup for MLLMs. Specifically, input is formatted as: \texttt{[Video] [Context] [Video]} with identical video, where the system prompt explicitly informs the model of the redundancy of input video.  As shown in \cref{tab:video_text_ordering} (b), Gemini achieves a notable 2.1\% performance gain with two repeated videos as input. This is surprising, as autoregressive MLLMs theoretically have the capability to revisit the video multiple times during answer generation, even if the video is only presented once. This finding suggests that, despite its remarkable capabilities, a powerful MLLM like Gemini still has suboptimal reasoning processes for Video QA.

\section{Visualization Results}
\label{sec:supp_visualization_results}
In this section, we present more qualitative results, including more examples of \bench, further error analysis case studies, examples of Chain-of-Thought promptings, and additional cognitive maps.

\subsection{\bench Examples}
\begin{figure*}[htbp] 
    \centering
    \includegraphics[width=0.88\textwidth]{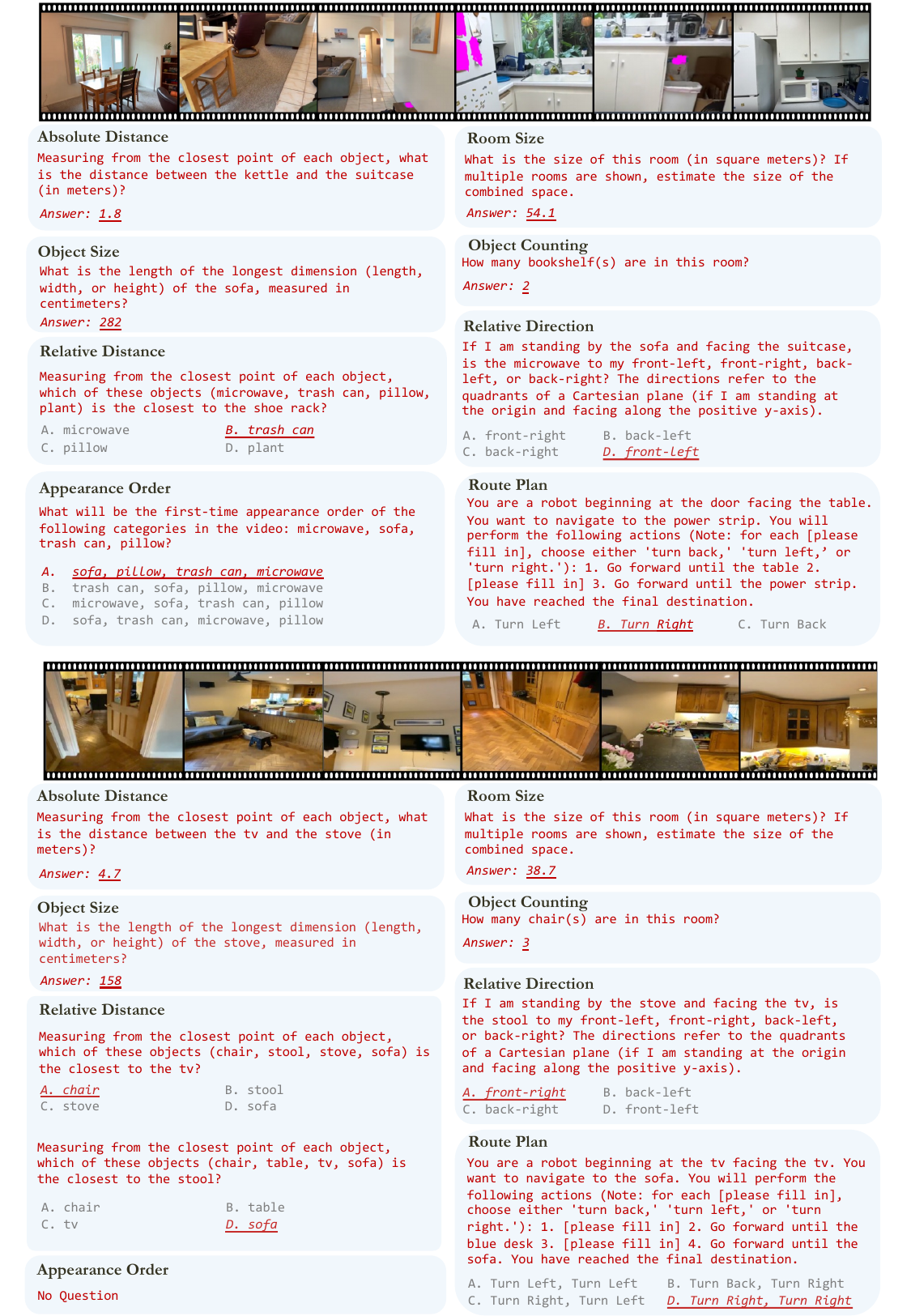} 
    \caption{\textbf{\bench Examples (Part 1).}}
    \label{fig:supp-vsi-sample_part1}
\end{figure*}

\begin{figure*}[htbp] 
    \centering
    \includegraphics[width=0.88\textwidth]{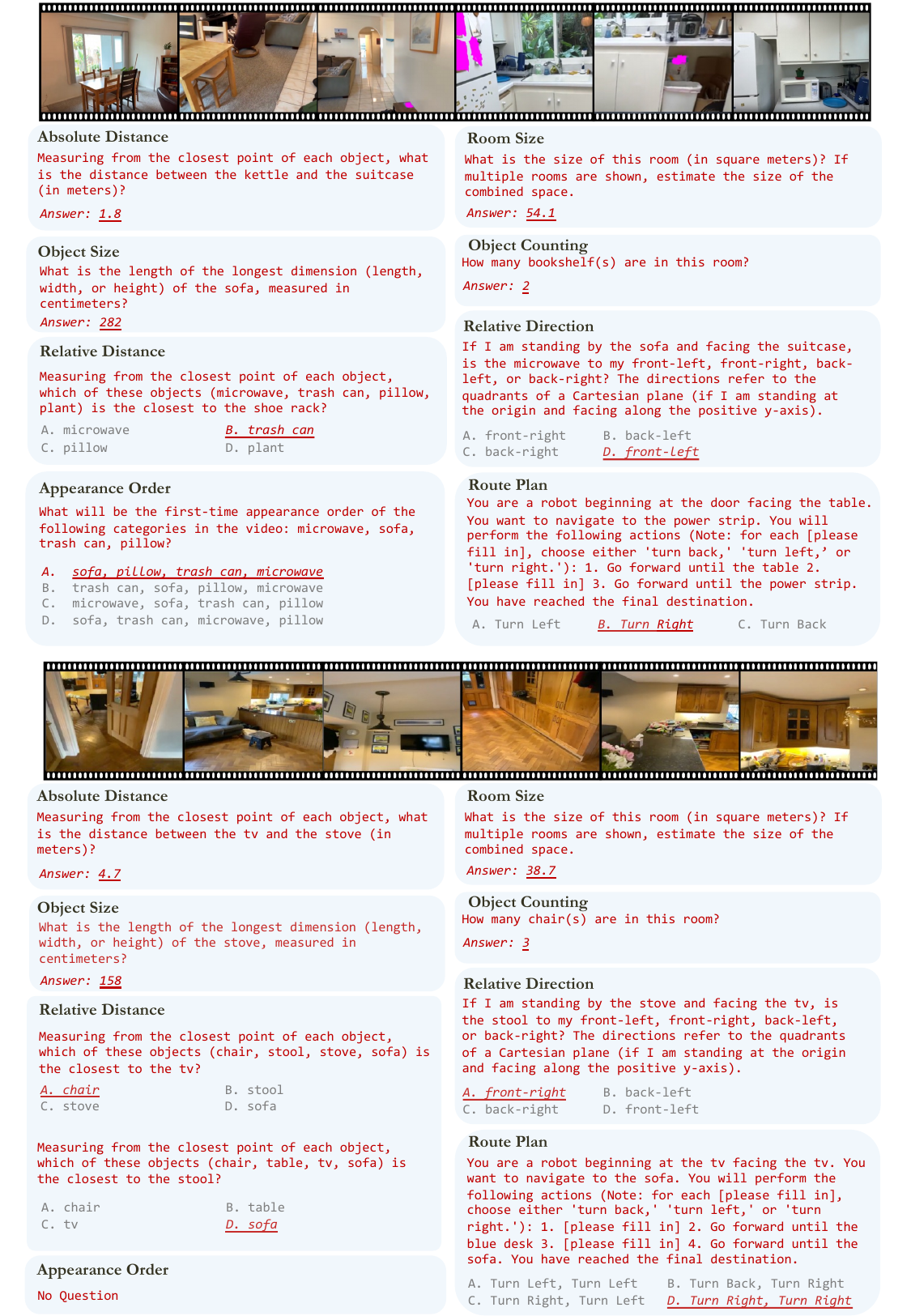} 
    \caption{\textbf{\bench Examples (Part 2).}}
    \label{fig:supp-vsi-sample_part2}
\end{figure*}

In \cref{fig:supp-vsi-sample_part1} and \cref{fig:supp-vsi-sample_part2}, we provide more examples from \bench to illustrate the structure and format of tasks, questions, and answers. 

\subsection{Error Analysis Examples}
\label{sec:supp_error_analysis}
\begin{figure*}[htbp] 
    \centering
    \includegraphics[width=1\textwidth]{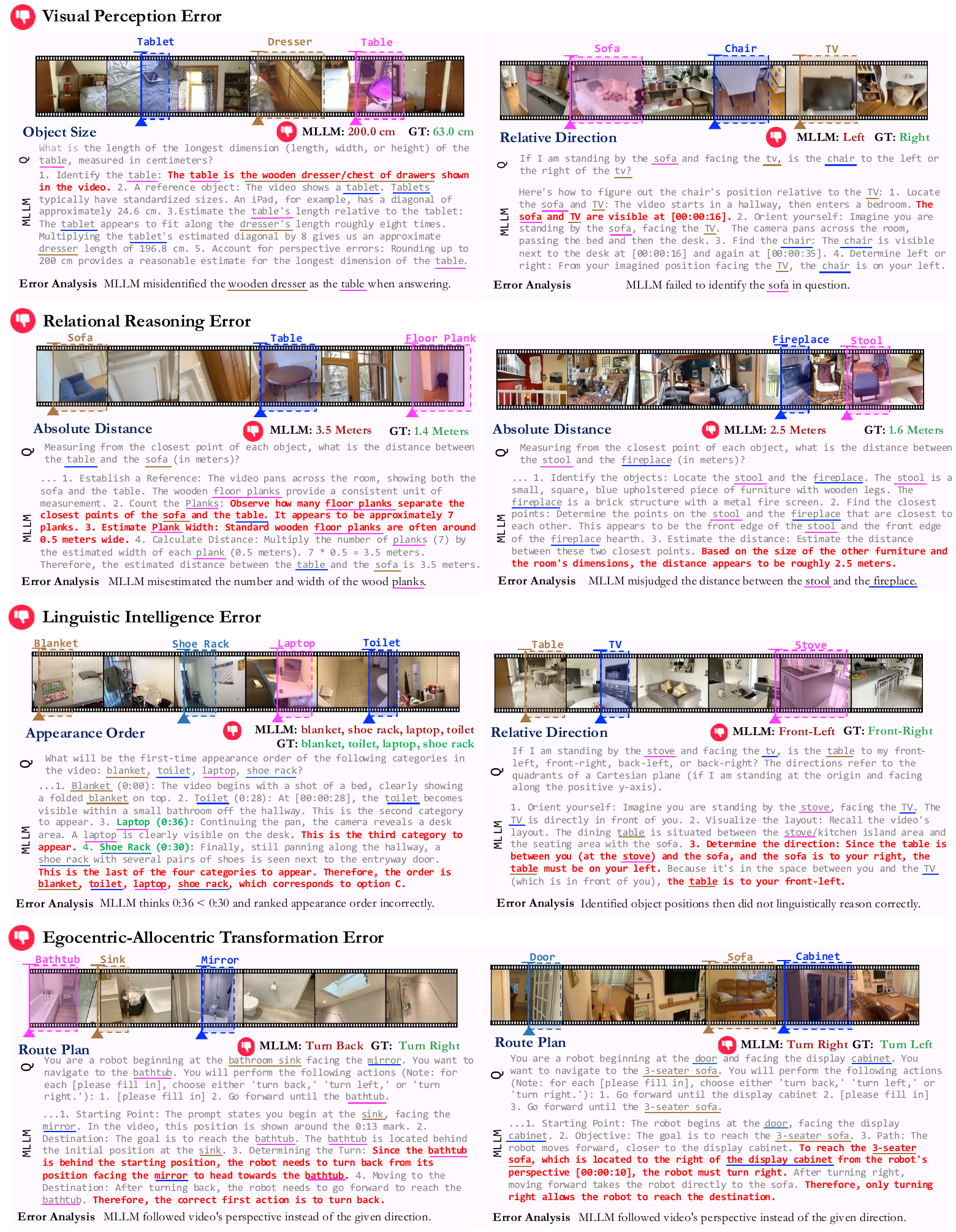} 
    \caption{\textbf{Additional Error Analysis Examples.}}
    \label{fig:supp_error_analysis}
\end{figure*}

In \cref{fig:supp_error_analysis}, we present more case studies for our human-conducted error analysis on \bench. In the error analysis, we identify the categorized error types and highlight the relevant parts of the explanation.

\begin{table}[t]
    \small
    \centering
    \setlength\tabcolsep{4pt} 
    \renewcommand{\arraystretch}{1.0}
    \scalebox{1}{
    \begin{tabular}{r|c}
    Methods & \# of Frames \\
    \hline
    \rowcolor{navyblue!5}
    \multicolumn{1}{l|}{\textcolor{black}{\textit{Proprietary Models (API)}}} & \\
    GPT-4o & 16 \\
    Gemini-1.5 Flash & - \\
    Gemini-1.5 Pro & - \\
    \hline
    \rowcolor{navyblue!5}
    \multicolumn{1}{l|}{\textcolor{black}{\textit{Open-source Models}}} & \\
    InternVL2-2B & 32 \\
    InternVL2-8B & 32 \\
    InternVL2-40B & 32 \\
    LongVILA-8B & 32 \\
    VILA-1.5-8B & 32 \\
    VILA-1.5-40B & 32 \\
    LongVA-7B & 32 \\
    LLaVA-Video-7B & 32 \\
    LLaVA-Video-72B & 32 \\
    LLaVA-OneVision-0.5B & 32 \\
    LLaVA-OneVision-7B & 32 \\
    LLaVA-OneVision-72B & 32 \\
    \end{tabular}}
    \caption{\textbf{Number of frames used in evaluation.}}
    \label{tab:number_of_frames_setup}
\end{table}

\subsection{Linguistic Prompting Examples}
\begin{figure*}[htbp] 
    \centering
    \includegraphics[width=1\textwidth]{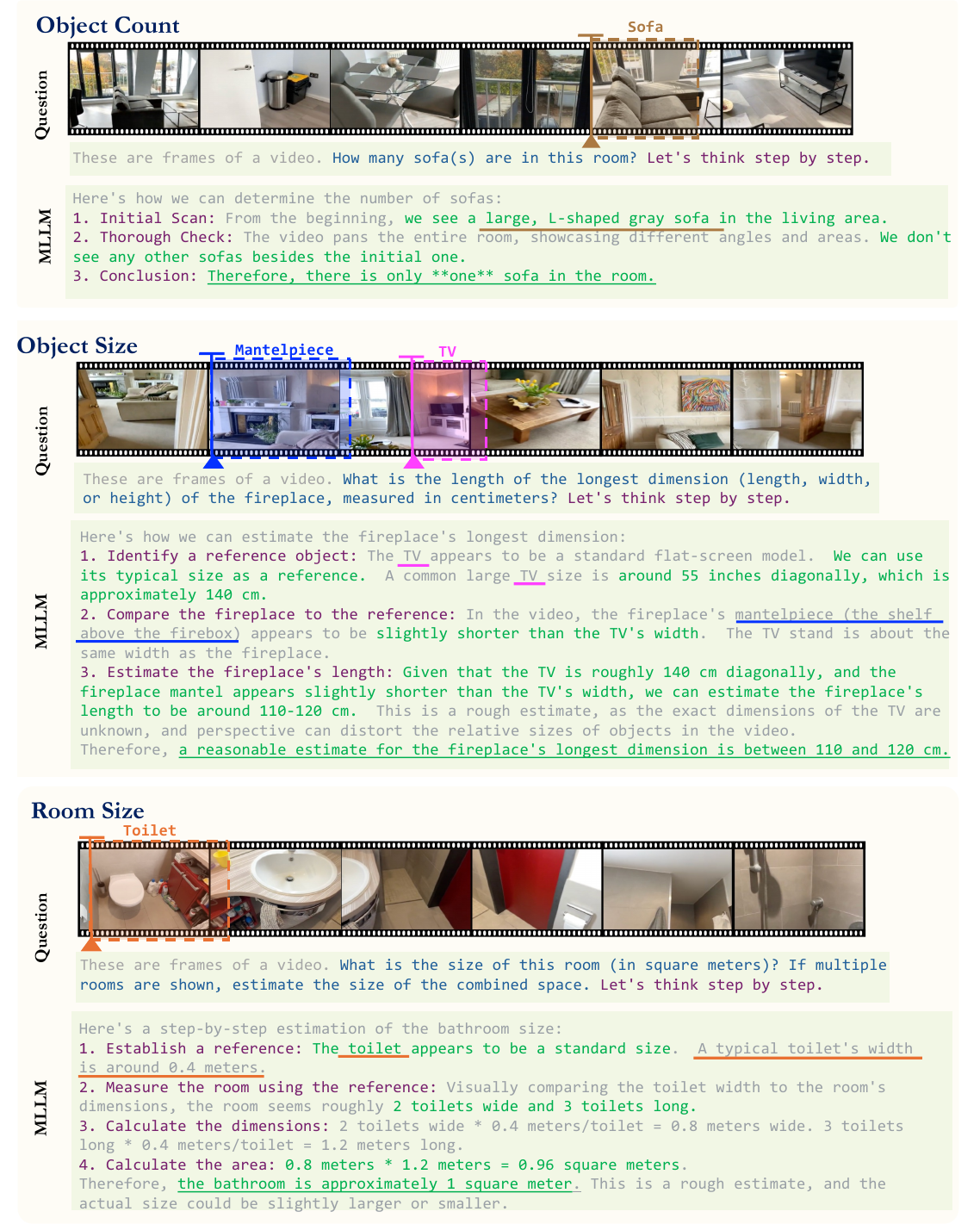} 
    \caption{\textbf{Zero-Shot CoT Examples.}}
    \label{fig:supp_zero_shot_cot}
\end{figure*}

\begin{figure*}[htbp] 
    \centering
    \includegraphics[width=1\textwidth]{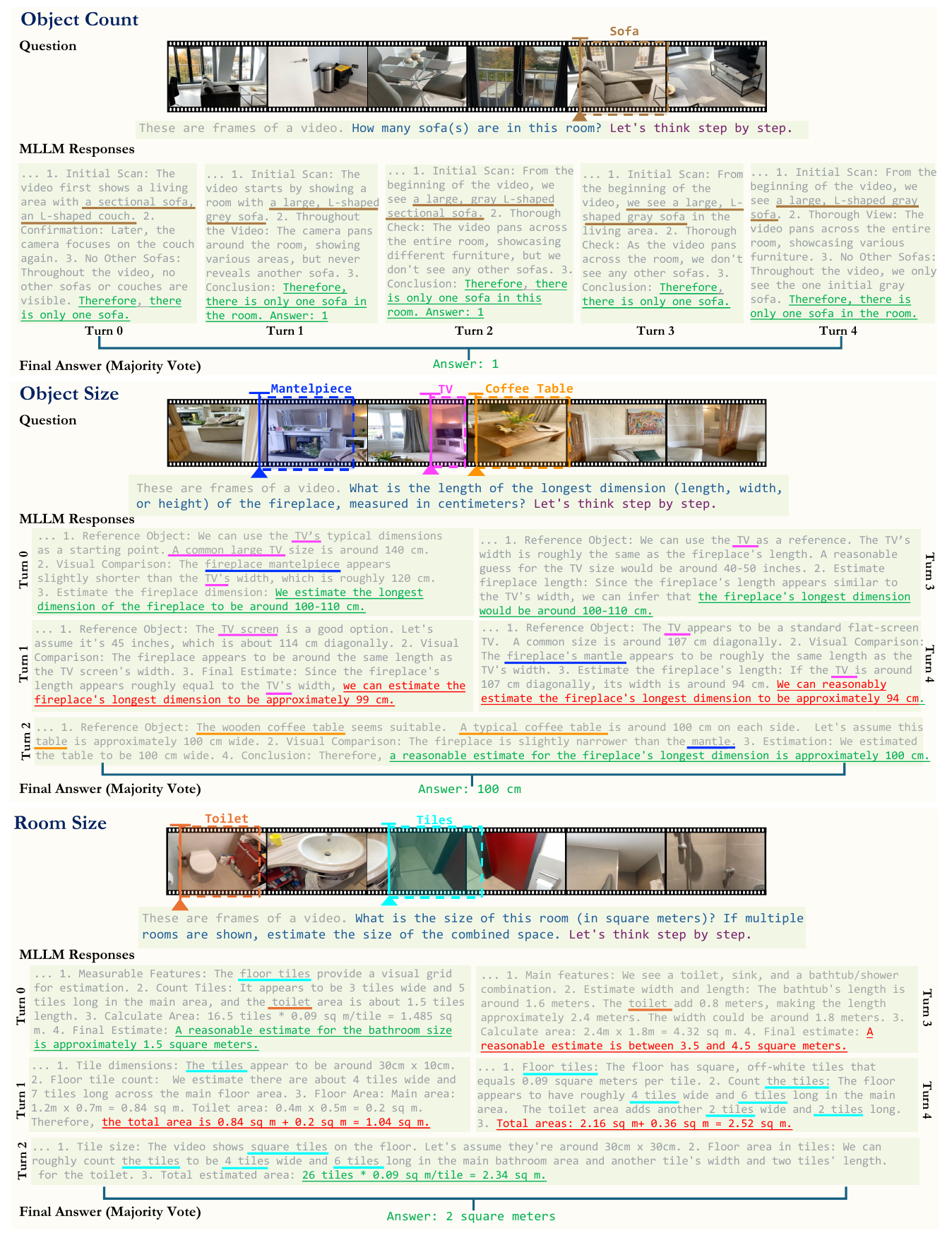} 
    \caption{\textbf{Self-Consistency w/ CoT Examples.}}
    
    \label{fig:supp_zero_shot_self-consistency}
\end{figure*}

\begin{figure*}[htbp] 
    \centering
    \includegraphics[width=1\textwidth]{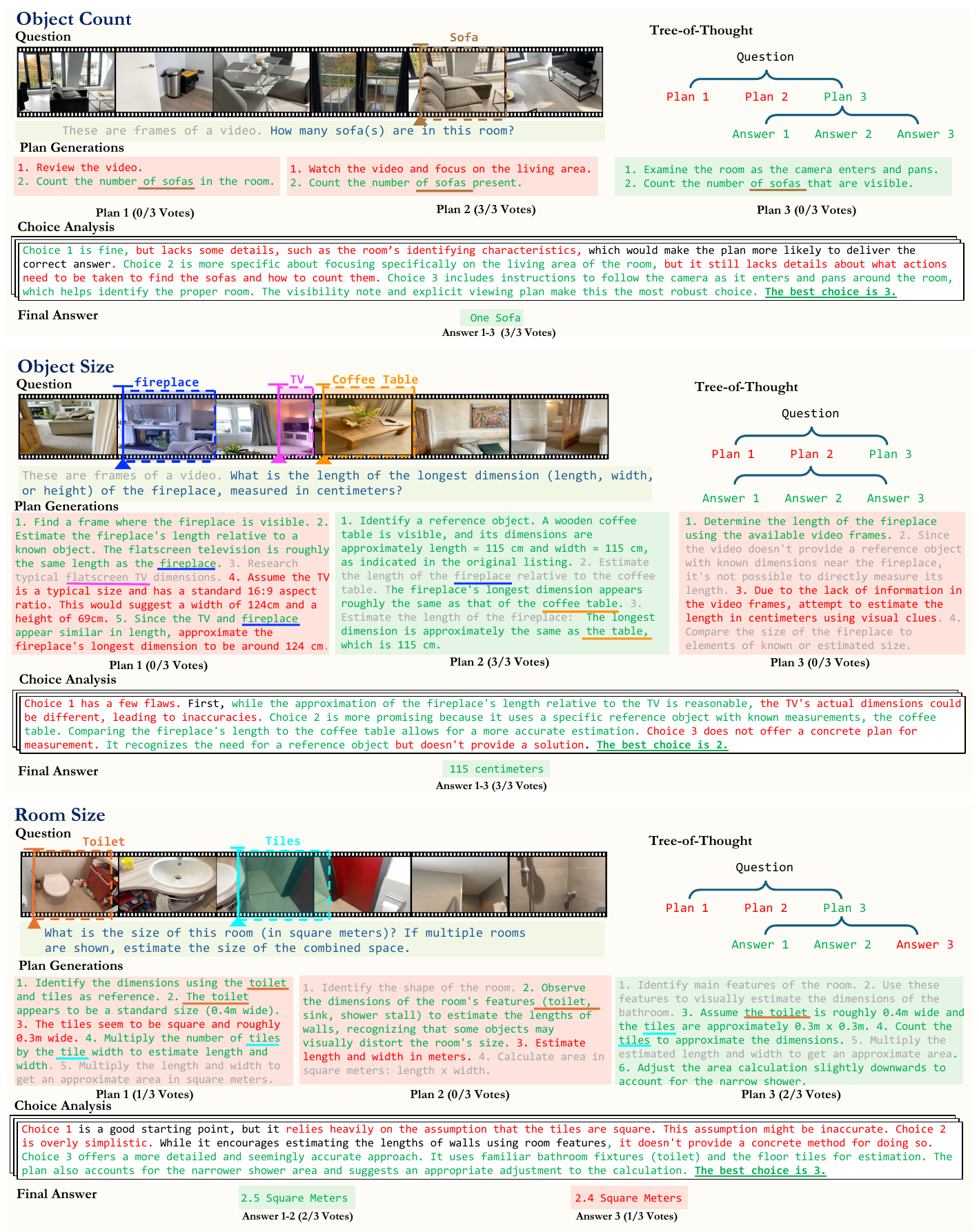} 
    \caption{\textbf{Tree-of-Thought Examples.}}
    
    \label{fig:supp-tot}
\end{figure*}

We provide examples for the three CoT prompting methods discussed in \cref{sec:cot_limits_visuospatial} to illustrate their concrete reasoning procedure in detail. 
We include examples of three selected tasks: object count, object size, and room size. 
For Zero-Shot Chain of Thought, as shown in \cref{fig:supp_zero_shot_cot}, we highlight each step of the MLLM's reasoning process to offer insights into how it arrives at its final decision. 
For Self-Consistency w/ CoT, as illustrated in \cref{fig:supp_zero_shot_self-consistency}, each example is paired with five independent responses. The final answer is then determined by a majority vote. 
For Tree-of-Thought, \cref{fig:supp-tot} details how each depth of the decision tree is reached. At the first depth, the MLLM generates three potential plans and conducts a choice analysis to select the optimal plan. At the second and final depth, the selected plan is used to generate three potential answers, with the final output determined through a majority vote.

\subsection{Cognitive Map Examples}
\begin{figure*}[htbp] 
    \centering
    \includegraphics[width=1\textwidth]{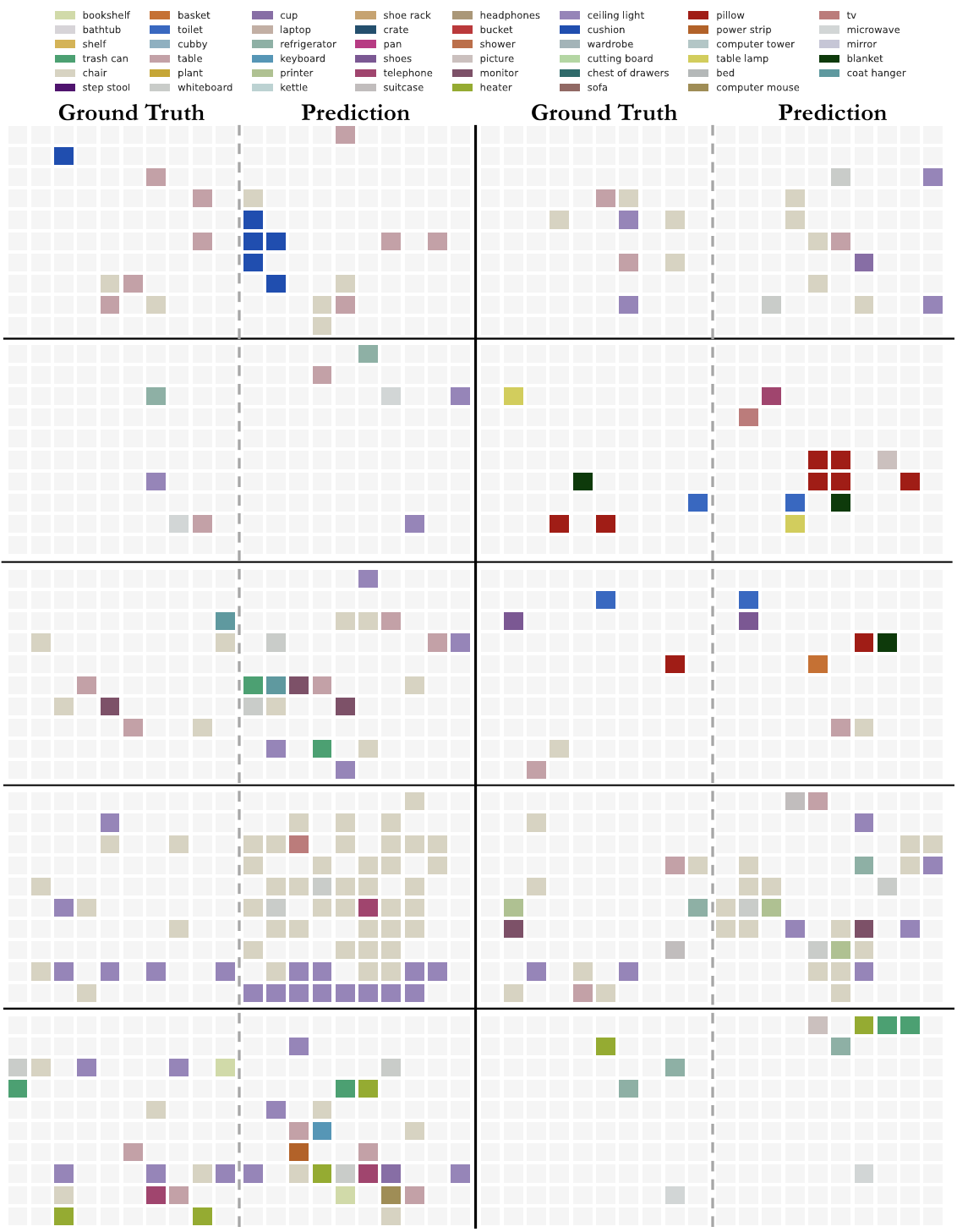} 
    \caption{\textbf{Additional predicted cognitive map examples.}}
    \label{fig:supp_cog_map}
\end{figure*}

In \cref{fig:supp_cog_map}, we include 10 additional cognitive maps and pair each prediction with its corresponding ground truth map to provide insight into the alignment between predicted and ground truth layouts.

\begin{table*}[ht!]
    \renewcommand{\arraystretch}{1.1}
    \small
    \centering
    \begin{tabular}{c|c|c|l}
         & Models & QA. Type & Prompt \\
        \hline
        {\small\texttt{Pre-Prompt}} & - & - & \textit{These are frames of a video.} \\
        \hline
        \multirow{4}{*}{{\small\texttt{Post-Prompt}}} & \multirow{2}{*}{Open-source Models} & NA & \textit{Please answer the question using a single word or phrase.} \\
        & & MCA & \textit{Answer with the option's letter from the given choices directly.} \\
        & \multirow{2}{*}{Proprietary Models} & NA & \textit{Do not respond with anything other than a single number!} \\
        & & MCA & \textit{Answer with the option's letter from the given choices directly.} \\
    \end{tabular}
    \caption{\textbf{Prompts used in evaluation.} NA and MAC indicates questions with \textit{Numerical Answer} and \textit{Multiple Choice Answer} respectively.}
    \label{tab:evaluation_prompt}
\end{table*}

\begin{table*}[t]
    \centering
    \fontsize{4.6pt}{4.4pt}\selectfont
    \setlength\tabcolsep{4pt}
    \renewcommand{\arraystretch}{1.2}
    \scalebox{2.1}{
    \begin{tabular}{r|c|cccccccc}
    & &
    \rotatebox{75}{Obj. Count} &
    \rotatebox{75}{Abs. Dist.} &
    \rotatebox{75}{Obj. Size} & 
    \rotatebox{75}{Room Size} &
    \rotatebox{75}{Rel. Dist.} &
    \rotatebox{75}{Rel. Dir.} &
    \rotatebox{75}{Route Plan} &
    \rotatebox{75}{Appr. Order} \\
    Methods & Avg. & \multicolumn{4}{c}{\cellcolor{orange!10}Numerical Answer} & \multicolumn{4}{c}{\cellcolor{yellow!10}Multiple-Choice Answer} \\
    \hline
    \rowcolor{navyblue!5}
    \multicolumn{1}{l|}{\textcolor{black}{\textit{Proprietary Models (API)}}} & & & & & & & & & \\
    Gemini-1.5 Flash & 41.6 & 49.1 & 30.3 & 52.7 & 53.7 &	37.1 & 40.8 &	31.4 &	37.1 \\
    Gemini-1.5 Pro & 44.9 &	55.1 &	30.3 &	63.1 &	43.3 & 50.0 & 45.9 & 35.7 & 35.7 \\
    \hline
    \rowcolor{navyblue!5}
    \multicolumn{1}{l|}{\textcolor{black}{\textit{Open-source Models}}} & & & & & & & & & \\
    InternVL2-2B & 27.0 & 22.4 &	24.9 &	21.1 &	34.1 &	32.9 &	43.5 &	30.0 &	7.1 \\
    InternVL2-8B & 34.1& 22.6& 28.3& 47.6& 39.6& 35.7& 30.4& 30.0& 38.6 \\
    InternVL2-40B &35.5& 34.4& 26.9& 45.6& 31.3& 41.4& 31.7& 32.9& 40.0 \\
    LongVILA-8B & 21.0& 28.7& 8.6& 16.3& 0.0& 28.6& 30.5& 31.4& 24.3 \\
    VILA-1.5-8B & 28.4& 17.3& 21.6& 49.9& 18.6& 31.4& 34.4& 30.0& 24.3 \\
    VILA-1.5-40B & 30.8& 21.4& 24.4& 48.3& 21.9& 40.0& 25.0& 30.0& 35.7 \\
    LongVA-7B & 29.0& 38.1& 16.9& 38.1& 21.7& 32.9& 42.8& 25.7& 15.7 \\
    LLaVA-Video-7B & 34.9& 47.9& 13.4& 46.7& 23.9& 42.9& 41.9& 32.9& 30.0 \\
    LLaVA-Video-72B & 40.5& 48.3& 22.6& 56.7& 34.6& 41.4& 36.5& 35.7& 48.6 \\
    LLaVA-OneVision-0.5B & 27.6& 45.1& 27.9& 14.7& 27.9& 28.6& 37.0& 34.3& 5.7 \\
    LLaVA-OneVision-7B & 32.1& 46.9& 19.9& 46.9& 12.1& 41.4& 35.1& 30.0& 24.3 \\
    LLaVA-OneVision-72B & 39.6& 42.7& 23.7& 56.7& 36.9& 41.4& 39.5& 31.4& 44.3 \\
    \end{tabular}
    }
\captionsetup{font={small}}
    \caption{\textbf{Evaluation results on \bench 560 samples subset.}}
\label{tab:560_subset_results}
\end{table*}

\begin{table*}[t]
    \centering
    \fontsize{4.6pt}{4.4pt}\selectfont
    \setlength\tabcolsep{4pt}
    \renewcommand{\arraystretch}{1.2}
    \scalebox{2.1}{
    \begin{tabular}{r|c|cccccccc}
    & &
    \rotatebox{75}{Obj. Count} &
    \rotatebox{75}{Abs. Dist.} &
    \rotatebox{75}{Obj. Size} & 
    \rotatebox{75}{Room Size} &
    \rotatebox{75}{Rel. Dist.} &
    \rotatebox{75}{Rel. Dir.} &
    \rotatebox{75}{Route Plan} &
    \rotatebox{75}{Appr. Order} \\
    Methods & Avg. & \multicolumn{4}{c}{\cellcolor{orange!10}Numerical Answer} & \multicolumn{4}{c}{\cellcolor{yellow!10}Multiple-Choice Answer} \\
    \hline
    \rowcolor{navyblue!5}
    \multicolumn{1}{l|}{\textcolor{black}{\textit{Proprietary Models (API)}}} & & & & & & & & & \\
    GPT-4o & 35.6 & 36.2 & 4.6 & 47.2 & 40.4 & 40.0 & 46.2 & 32.0 & 38.0 \\
    Gemini-1.5 Flash & 45.7 & 50.8 & 33.6 & 56.5 & 45.2 & 48.0 & 39.8 & 32.7 & 59.2 \\
    Gemini-1.5 Pro & 48.8 & 49.6 & 28.8 & 58.6 & 49.4 & 46.0 & 48.1 & 42.0 & 68.0 \\
    Gemini-2.0 Flash & 45.4 & 52.4 & 30.6 & 66.7 & 31.8 & 56.0 & 46.3 & 24.5 & 55.1 \\
    \hline
    \rowcolor{navyblue!5}
    \multicolumn{1}{l|}{\textcolor{black}{\textit{Open-source Models}}} & & & & & & & & & \\
    InternVL2-2B & 25.5 & 30.6 & 20.4 & 26.0 & 29.6 & 28.0 & 39.2 & 28.0 & 2.0 \\
    InternVL2-8B & 32.9 & 26.4 & 25.4 & 43.8 & 41.6 & 30.0 & 32.2 & 20.0 & 44.0 \\
    InternVL2-40B & 37.6 & 40.8 & 23.8 & 48.0 & 26.0 & 46.0 & 30.1 & 42.0 & 44.0 \\
    LongVILA-8B & 19.1 & 23.4 & 10.8 & 11.4 & 0.0 & 20.0 & 33.1 & 28.0 & 26.0 \\
    VILA-1.5-8B & 31.4 & 12.2 & 23.4 & 51.4 & 18.6 & 36.0 & 41.5 & 42.0 & 26.0 \\
    VILA-1.5-40B & 32.3 & 14.6 & 21.0 & 48.0 & 20.6 & 42.0 & 22.0 & 40.0 & 50.0 \\
    LongVA-7B & 31.8 & 41.2 & 17.4 & 39.6 & 25.4 & 30.0 & 52.8 & 34.0 & 14.0 \\
    LLaVA-Video-7B & 35.7 & 49.0 & 12.8 & 48.6 & 21.4 & 40.0 & 43.5 & 34.0 & 36.0 \\
    LLaVA-Video-72B & 39.3 & 41.4 & 26.6 & 55.6 & 31.6 & 36.0 & 25.6 & 42.0 & 56.0 \\
    LLaVA-OneVision-0.5B & 27.7 & 44.0 & 23.0 & 18.8 & 28.4 & 30.0 & 33.4 & 36.0 & 8.0 \\
    LLaVA-OneVision-7B & 33.8 & 48.2 & 22.0 & 44.4 & 14.0 & 44.0 & 31.9 & 34.0 & 32.0 \\
    LLaVA-OneVision-72B & 41.6 & 38.0 & 31.6 & 54.4 & 35.2 & 44.0 & 39.7 & 32.0 & 58.0 \\
    \end{tabular}
    }
\captionsetup{font={small}}
    \caption{\textbf{Complete \benchtiny evaluation results.}}
\label{tab:complete_subset_results}
\end{table*}

\begin{table*}[ht!]
    \centering
    \fontsize{4.6pt}{4.4pt}\selectfont
    \setlength\tabcolsep{4pt} 
    \renewcommand{\arraystretch}{1.2} 
    \scalebox{2.1}{
    \begin{tabular}{r|c|cccccccc}
    & &
    \rotatebox{75}{Obj. Count} &
    \rotatebox{75}{Abs. Dist.} &
    \rotatebox{75}{Obj. Size} & 
    \rotatebox{75}{Room Size} &
    \rotatebox{75}{Rel. Dist.} &
    \rotatebox{75}{Rel. Dir.} &
    \rotatebox{75}{Route Plan} &
    \rotatebox{75}{Appr. Order} \\
    Methods & Avg. & \multicolumn{4}{c}{\cellcolor{orange!10}Numerical Answer} & \multicolumn{4}{c}{\cellcolor{yellow!10}Multiple-Choice Answer} \\
    \hline
    \rowcolor{navyblue!5}
    \multicolumn{1}{l|}{\textcolor{black}{\textit{Proprietary Models (API)}}} & & & & & & & & & \\
    GPT-4o & 14.5 & 0.1 & 5.2 & 36.7 & 0.0 & 10.8 & 23.2 & 26.9 & 13.1 \\
    Gemini-1.5 Flash & 19.9 & 25.0 & 30.3 & 52.5 & 0.0 & 0.0 & 21.2 & 29.9 & 0.2 \\
    Gemini-1.5 Pro & 32.3 & 30.6 & 11.5 & 51.5 & 33.1 & 33.8 & 44.6 & 33.5 & 20.2 \\
    \hline
    \rowcolor{navyblue!5}
    \multicolumn{1}{l|}{\textcolor{black}{\textit{Open-source Models}}} & & & & & & & & & \\
    InternVL2-2B & 17.8 & 5.4 & 23.7 & 9.2 & 0.0 & 26.9 & 41.2 & 27.9 & 7.9 \\
    InternVL2-8B & 27.6 & 31.9 & 26.8 & 38.3 & 0.7 & 27.1 & 39.2 & 33.0 & 23.6 \\
    InternVL2-40B & 24.4 & 5.4 & 29.1 & 39.2 & 0.7 & 30.3 & 37.7 & 27.9 & 24.7 \\
    LongVILA-8B & 20.2 & 47.4 & 12.6 & 8.7 & 0.6 & 24.3 & 27.0 & 27.4 & 13.9 \\
    VILA-1.5-8B & 21.5 & 7.4 & 7.6 & 45.7 & 0.0 & 25.4 & 39.1 & 29.4 & 17.6 \\
    VILA-1.5-40B & 25.5 & 5.3 & 27.6 & 46.5 & 0.7 & 30.2 & 37.1 & 31.5 & 25.0 \\
    LongVA-7B & 21.9 & 5.1 & 18.1 & 27.4 & 26.1 & 23.4 & 39.8 & 26.9 & 8.7 \\
    LLaVA-Video-7B & 25.2 & 14.8 & 14.6 & 32.5 & 26.1 & 26.8 & 45.0 & 33.0 & 8.5 \\
    LLaVA-Video-72B & 29.1 & 19.0 & 25.4 & 46.3 & 26.1 & 29.0 & 38.8 & 33.0 & 15.5 \\
    LLaVA-OneVision-0.5B & 28.6 & 38.4 & 30.1 & 32.0 & 24.3 & 22.0 & 41.8 & 34.5 & 5.4 \\
    LLaVA-OneVision-7B & 25.3 & 13.8 & 8.5 & 45.5 & 26.1 & 28.6 & 41.2 & 27.9 & 11.1 \\ 
    LLaVA-OneVision-72B & 28.9 & 8.2 & 23.8 & 54.1 & 26.1 & 30.4 & 38.1 & 33.0 & 17.1 \\
    \end{tabular}}
    \caption{\textbf{Complete blind evaluation results.}}
    \label{tab:complete_blind_evaluation_results}
    \vspace{-0.4cm}
\end{table*}

\begin{table*}[ht!]
    \centering
    \fontsize{4.6pt}{4.4pt}\selectfont
    \setlength\tabcolsep{4pt}
    \renewcommand{\arraystretch}{1.2} 
    \scalebox{2.1}{
    \begin{tabular}{r|c|cccccccc}
    & &
    
    \rotatebox{75}{Obj. Count} &
    \rotatebox{75}{Abs. Dist.} &
    \rotatebox{75}{Obj. Size} & 
    \rotatebox{75}{Room Size} &
    
    \rotatebox{75}{Rel. Dist.} &
    \rotatebox{75}{Rel. Dir.} &
    \rotatebox{75}{Route Plan} &
    \rotatebox{75}{Appr. Order} \\
    Methods & Avg. & \multicolumn{4}{c}{\cellcolor{orange!10}Numerical Answer} & \multicolumn{4}{c}{\cellcolor{yellow!10}Multiple-Choice Answer} \\
    \hline

    \rowcolor{navyblue!5}
    \multicolumn{1}{l|}{\textcolor{black}{\textit{Proprietary Models (API)}}} & & & & & & & & & \\
    GPT-4o & 19.5 & 46.1 & 0.1 & 7.1 & 38.2 & 26.2 & 18.0 & 4.6 & 15.4 \\
    Gemini-1.5 Flash & 22.2 & 24.9 & 0.5 & 1.0 & 54.4 & 37.7 & 19.9 & 1.5 & 37.7 \\
    Gemini-1.5 Pro & 13.0 & 25.5 & 19.5 & 12.6 & 10.6 & 17.5 & 1.7 & 2.5 & 14.4 \\
    \hline
    \rowcolor{navyblue!5}
    \multicolumn{1}{l|}{\textcolor{black}{\textit{Open-source Models}}} & & & & & & & & & \\
    InternVL2-2B & 8.7 & 20.3 & 0.3 & 10.8 & 29.2 & 5.2 & 2.9 & 2.5 & -1.6 \\
    InternVL2-8B & 9.9 & -0.6 & 2.2 & 10.6 & 43.5 & 10.9 & -5.8 & -4.1 & 22.8 \\
    InternVL2-40B & 12.6 & 35.9 & -2.9 & 9.0 & 26.8 & 17.3 & -5.0 & 9.9 & 20.0 \\
    LongVILA-8B & 1.4 & -18.2 & -3.5 & 7.9 & -0.6 & 5.3 & 3.7 & 5.1 & 11.5 \\
    VILA-1.5-8B & 7.3 & 10.0 & 14.2 & 4.6 & 18.8 & 6.7 & -4.4 & 1.5 & 7.2 \\
    VILA-1.5-40B & 5.7 & 17.1 & -2.8 & 2.2 & 22.0 & 10.4 & -11.4 & 0.0 & 7.9 \\
    LongVA-7B & 7.2 & 32.9 & -1.5 & 11.5 & -3.9 & 9.7 & 3.5 & -1.5 & 7.1 \\
    LLaVA-Video-7B & 10.5 & 33.8 & -0.6 & 15.2 & -1.9 & 16.7 & -2.7 & 1.0 & 22.1 \\
    LLaVA-Video-72B & 11.7 & 29.9 & -2.6 & 11.1 & 9.2 & 13.3 & -2.0 & 2.0 & 33.0 \\
    LLaVA-OneVision-0.5B & -0.5 & 7.8 & -1.7 & -16.6 & 4.0 & 6.9 & -5.0 & 0.0 & 0.3 \\
    LLaVA-OneVision-7B & 7.0 & 33.9 & 11.7 & 1.9 & -13.9 & 13.9 & -6.0 & 1.5 & 13.3 \\
    LLaVA-OneVision-72B & 11.4 & 35.4 & 0.1 & 3.5 & 11.4 & 12.1 & 1.8 & -0.5 & 27.4 \\
    \end{tabular}}
    \caption{\textbf{Results of Vision Enabled $-$ Vision Disabled.}}
    \label{tab:vision_enabled_disable}
    \vspace{-0.4cm}
\end{table*}

\end{document}